\documentclass[11pt, a4paper, copyright, gdm]{google} 

\usepackage[authoryear, sort&compress, round]{natbib}
\usepackage{todonotes}
\usepackage{svg}
\usepackage{wrapfig}
\usepackage{cleveref}
\usepackage[most]{tcolorbox}

\usepackage{listings}
\usepackage{xspace}
\usepackage{graphicx}
\usepackage{xcolor} 

\newtcolorbox[auto counter]{promptbox}[2][]{
    breakable,                  
    colback=gray!5!white,       
    colframe=gray!60!black,     
    fontupper=\small,           
    arc=4pt,                    
    boxrule=0.5pt,              
    left=6pt, right=6pt,        
    top=6pt, bottom=6pt,
    title={\textbf{Prompt \thetcbcounter: #2}}, 
    label={#1}                  
}


\uselogo{}

\newcommand{\tempname}{SkillSmith\xspace}

\title{SkillSmith: Learning to Compose Parametric Skills and Textual Knowledge}

\correspondingauthor{ldery@google.com, atjandr@google.com}

\author[*,1]{Lucio M. Dery}
\author[*,1]{Benedict Aaron Tjandra}
\author[1]{Siavash Samiei}
\author[1]{Adhiguna Kuncoro}
\author[1]{Zohar Yahav}
\author[1]{Jiajun Shen}
\author[1]{Arthur Szlam}

\affil[1]{\thepa{}{}}
\affil[*]{Equal contributions}

\begin{abstract}
We instantiate parametric learning via prefix-tuning and augment an LLM to ingest both prefix weights and rich textual data which capture  relationships to a target capability. Our augmented LLM, which we call \tempname, synthesizes these inputs to perform instruction-steered parametric synthesis, directly outputting new prefix weights that manifest the target skill. 
We demonstrate that our approach significantly outperforms both text-only and weight-space-only baselines, unlocking performance gains that are out of reach for uni-modal (text-only or weight-only) adaptations.
\end{abstract}

\begin{document}

\maketitle

\section{Introduction}

\looseness-1 LLMs \citep{gpt4, gemini25, claude3, qwen25, gemmateam2025gemma3technicalreport} have evolved from static conversational interfaces into drivers of agentic systems capable of solving problems that require complex multi-step reasoning \citep{metr, sima2, imo2025}. Central to the efficacy of these systems is their ability to learn and adapt from past experiences. Currently, this adaptation is driven by two powerful, yet largely siloed mechanisms. The first is the synthesis of text-based knowledge, where agents use natural language---such as self-reflection \citep{shinn2023reflexion, self-refine, renze2024selfreflection}, structured memory \citep{mem0, a-mem, open-source-phoenix}, or prompt generation \citep{promptagent, gepa}---to guide future reasoning and planning. The second involves building parametric skill libraries via parameter-efficient fine-tuning (PEFT) \citep{shekar2025adaptive} for consolidating learned behaviors into modular weights which can be retrieved and used independently or merged with other skills to efficiently address recurring sub-goals \citep{huang2023lorahub, pfeiffer2020adapterhub}.

\looseness-1 Despite burgeoning activity in agentic research \citep{sun2025training, zhang2025agentic, zhou2025memento}, to date, the community has largely treated text-based reasoning and parametric skill acquisition as orthogonal pursuits. We believe this separation limits the potential of agentic systems. Empowering agents to seamlessly reason over both weight-space and text could unlock compositional generalization \citep{dziri2023faith, keysers2019measuring, lake2018generalization} across parametric skills, with the added advantage that the axes of composition and generalization can be flexibly steered via instruction text. As a concrete example, consider an agent that has developed a set of models (parametric skills) for solving various subtasks, alongside a rich, text-based strategy for deploying them. Suppose that through past user interactions, this agent trained a prefix-cache \citep{prefix-tuning} for translating English to Twi in one session, and compiled comprehensive notes on analyzing English legal documents in another. When encountering a novel task, e.g. \textit{analyzing a legal document written in Twi}, the agent reasons that the solution requires a combination of its prior skills: \textit{English-to-Twi translation}, \textit{Twi language modeling} and \textit{legal document analysis}. However, while the agent can articulate this transfer textually, there is currently no mechanism for it to leverage this reasoning along with the weight-space  \textit{translation skill} and \textit{Twi language skill} to directly synthesize the corresponding task weights for the new problem.

\begin{figure}[t!]
\begin{center}
 \includegraphics[width=\linewidth, keepaspectratio]{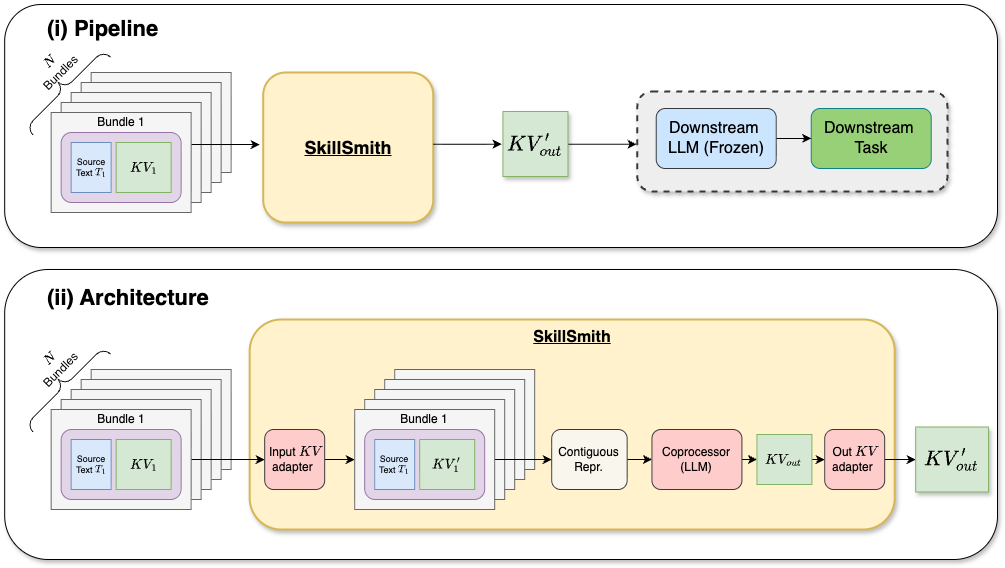}
\caption{High level \tempname pipeline (i) and architecture (ii). Source task bundles consisting of text and prefix-cache weights -- along with text rationales of how these task bundles can be synthesized to help the target task -- are fed as inputs to an augmented LLM to directly generate the prefix-cache that will be used to modulate a downstream LLM to the target task.}
\label{fig:method_image}  
\end{center}
\end{figure}

\looseness-1 This work takes initial steps toward the frictionless composition of text- and weight-based agentic artifacts. We propose to treat weight-space inputs as simply an additional modality that can be natively processed by an appropriately augmented pre-trained LLM. This augmented LLM -- which we term the \textbf{\tempname} -- is trained to simultaneously reason over both text and weight-space inputs, enabling it to directly generate task-specific parametric skills for downstream deployment (Figure \ref{fig:method_image}). Specifically, we first instantiate parametric skill learning as prefix-tuning \citep{prefix-tuning}. We then train \tempname to ingest (i) prefix-weights along with rich textual metadata about the tasks solved by the weights and (ii) extra text meta-data that captures the functional relationships between the (weights/) skills and a desired target capability. The target capability is specified via a high level text description or few shot exemplars and the output prefix-weights from \tempname are directly optimized on the target task in an end-to-end fashion as in \cite{liu2024deliberation}.

To demonstrate that \tempname flexibly composes text and weight-based agentic artifacts into target skills, we make the following contributions:
\begin{itemize}
    \item We provide a comprehensive study of baseline weight-space only merging methods for prefix-tuning. To the best of our knowledge, existing PEFT merging work has primarily focused on LoRA modules \citep{prabhakar2025lora, zhao2024merging, lora} and thus extensive base-lining of simple merging methods for prefix-tuning are absent.
    \item To study mixed modality composition at scale, we need robust datasets where ground-truth relationships between tasks are known. To this end, we contribute a data-generation workflow that we use to produce Composite SNI, a synthetic compositional generalization dataset. By systematically prompting Gemini 2.5 \citep{gemini25} to generate tasks that require the combined skill sets of two input tasks from the Super Natural Instructions (SNI) dataset \citep{wang2022super}, we create a setting where the ground-truth input tasks for a particular target capability are known. We will show that we can effectively bootstrap this synthetic data to jump-start learning text-and-weight composition in settings with a limited number of tasks.
    \item Using the 4B Gemma 3 model \citep{gemmateam2025gemma3technicalreport} for both \tempname  and downstream task-solving, we show that \tempname can outperform standard zero-shot, weight-space composition baselines on three datasets: Composite SNI, SNI and MMLU-ProX \citep{mmluprox}. Furthermore, we demonstrate that SkillSmith serves as a highly superior parameter initialization; fine-tuning the prefix-weights generated by our \tempname on the target task yields representations that either outperform or are competitive with all other methods studied, including the direct training of a prefix-cache initialized from in-context examples.
    \item Finally, we introduce a retriever-based approach for settings where the ground-truth mapping of set input tasks to target task capability are unknown. We show that even in this noisy setting, \tempname is able to extract and compose the relevant signal (if it exists) from the candidate source tasks and leverage that to generate prefix-weights that outperform baselines under equalized conditions.
\end{itemize}

Ultimately, \tempname demonstrates that language models can natively reason over (their own) modular weights just as they do with text, thus providing a blueprint for more holistic agentic architectures. Our results underscore that treating parameter-space as a readable, synthesizable modality provides a distinctly superior initialization for targeted adaptation, facilitating the scalable and effective composition of agentic skills.
\section{Related Work}

\noindent \textbf{Text-Based Agentic Adaptation.} Modern LLM-driven agents rely heavily on past experiences to solve complex problems. To date, this adaptation has been driven almost exclusively by synthesizing text-based knowledge. Agents leverage natural language to guide future planning through mechanisms such as self-reflection \citep{self-refine, renze2024selfreflection, shinn2023reflexion}, structured memory \citep{mem0, open-source-phoenix}, and prompt optimization \citep{gepa, promptagent}. While text provides a highly flexible and composable medium for articulating task transfer, it is strictly bound by inference-time context limits: whilst a task can be fully specified by listing all it's text examplars in-context, this does not scale.

\noindent \textbf{Parametric Skill Acquisition and Weight Merging.} Conversely, Parameter-Efficient Fine-Tuning (PEFT) adapts language models to downstream tasks by consolidating learned behaviors into efficient, modular weights \citep{peft-survey-2024}. Techniques such as adapters \citep{adapter-1, adapter-2, adapter-3}, LoRA \citep{lora}, and prompt or prefix tuning \citep{prompt-tuning, prefix-tuning} allow models to build parametric skill libraries. Prior work leverages these components to transfer learned behaviors via weight merging, using methods like SPoT \citep{spot}, ATTEMPT \citep{attempt}, AdapterHub \citep{pfeiffer2020adapterhub}, and LoRAHub \citep{huang2023lorahub}. However, these approaches rely on shallow arithmetic operations—such as simple averaging, concatenation, or routing—which are mathematically rigid and fail to leverage the rich semantic relationships between the tasks being merged.

\looseness-1 \noindent \textbf{Bridging the Modality Gap via Hyper-Networks.} Despite rapid progress, the community has largely treated text-based reasoning and parametric skill acquisition as orthogonal pursuits. By unifying these separated paradigms, we aim to achieve true compositional generalization \citep{dziri2023faith, keysers2019measuring, lake2018generalization}. Rather than performing shallow task arithmetic, we employ an LLM as a hyper-network \citep{shenaj2025lora}. Unlike existing hyper-network approaches, ours is designed to natively process weight-space inputs—specifically prefix KV caches, chosen because text snippets can naturally be translated into them via standard forward passes—alongside textual metadata. By treating parameter-space as a readable and synthesizable modality, \tempname translates an agent's textual reasoning into instruction-steered parametric synthesis.
\section{Bridging text and parameter modalities with \tempname}
\label{section:methodology}
We are interested in settings where an agent, as part of its learning procedure, is able to build up both text and weight-space artifacts to solve tasks. We assume that  the tasks the agent encounters over time are sufficiently related such that new tasks can be solved by synthesizing relevant information from past experiences. We concretize our problem setting below:

\subsection{Preliminaries}
\looseness-1 Let $\mathcal{T}_{src}$ be the set of all tasks that an agent has encountered before. We associate each task $T_i \in \mathcal{T}_{src}$ with a task bundle $b_i = (m_i, w_i)$ where $m_i$ is a learned PEFT module and $w_i$ represents task-relevant textual meta-data, such as in-context learning (ICL) demonstrations or task reflections \citep{shinn2023reflexion}. Though the modules $m_i$ can conceptually represent any PEFT method, this work focuses specifically on prefix-tuning. Therefore each $m_i$ is a Key-Value (K-V) prefix cache associated with a pre-specified, frozen base model $M_{\phi}$. The choice of prefix-tuning is motivated by the fact that text snippets can naturally be translated into K-V caches via forward passes through $M_{\phi}$. This provides \textit{a priori} confidence that it is feasible to bridge the modality gap by learning relationships between parametric K-V caches trained from scratch and those derived directly from text.

\subsection{Problem Statement}
\looseness-1 Consider an agent presented with a new task, $T_{new}$, to solve. Our goal is to produce a new PEFT module $m_{new}$ which can solve instances of $T_{new}$ that the agent may encounter. A baseline approach would be to train a completely new PEFT module using only data from $T_{new}$; however, this fails to leverage potentially useful information stored within the agent's historical experiences, $\mathcal{T}_{src}$. Hence, we assume that the agent can access a subset of the previously encountered source tasks, $\mathcal{T}_{src}\big[T_{new}\big] = \{T_1, \dots, T_N\}$, deemed relevant to constructing a solution to $T_{new}$\footnote{Finding this absolute ground-truth mapping a priori can be challenging, but our retrieval experiments demonstrate that simple embedding-based retrieval techniques provide an effective approximation.}. 

\looseness=-1 Uni-modal strategies exist to leverage $\mathcal{T}_{\text{src}}[T_{new}]$, but their siloed nature presents limitations:  
\begin{itemize}
    \item \textbf{Text-only strategies:} One can aggregate all available text metadata, i.e. ICL examples and reflections, from the source tasks and combine it directly with the text metadata of $T_{new}$ to serve as in-context data for solving $T_{new}$. While highly flexible, this approach is computationally expensive and constrained by inference-time context limits.
    \item \textbf{Weight-space strategies:} One can initialize from the relevant trained modules $\{m_i\}$ associated with the task in $\mathcal{T}_{src}[T_{new}]$ using standard weight-space merging techniques (e.g., averaging) with the option of subsequently tuning the merged weights on data from $T_{new}$. While inference-efficient, arithmetic merging ignores the rich, functional relationships articulated in the text metadata.  
\end{itemize}

We contend that these isolated approaches do not maximally reuse the computation spent to construct $\mathcal{T}_{src}$ and they miss out on performance gains that could be achieved by synthesizing information across interleaved modalities. We are thus motivated to answer: \textit{\textbf{How can we bridge these modalities, effectively combining both text-based and weight-space artifacts from $\mathcal{T}_{\text{src}}[T_{\text{new}}]$, to construct an $m_{\text{new}}$ that definitively outperforms purely uni-modal baselines?}}

\subsection{\tempname Architecture}
\label{subsec:_achitecture_}

\begin{figure}[b!]
\begin{center}
 \includegraphics[width=\linewidth, height=4.5cm, keepaspectratio]{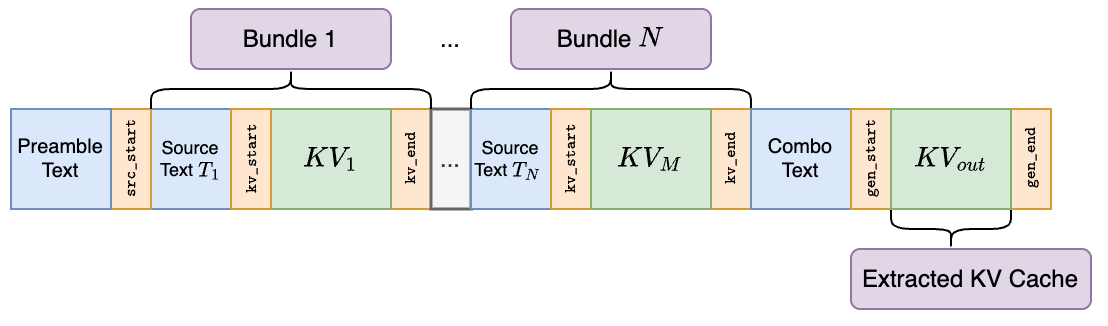}
\caption{Contiguous representation of task bundles to be processed by the co-processor.}
\label{fig:contiguous-repr} 
\end{center}
\end{figure}

Given the limitations of unimodal approaches as discussed, we introduce \tempname (\Cref{fig:method_image}), an augmented pre-trained transformer language model capable of synthesizing text and weight-space prefix-cache artifacts. To solve a new task $T_{new}$, \tempname composes existing related source task bundles $\mathcal{T}_{src}\big[T_{new}\big]$ alongside descriptive textual meta-data to generate $m_{new}$.

Specifically, given a set of $N$ task bundles $\{b_i\}_{N}$ as inputs, \tempname first
maps weight-space modules $\{m_i\}_{N}$ into the latent space of the language model by projection with an input K-V Adapter which we parameterize as a multi-layer perceptron (MLP). The structural sequence processed by the \tempname coprocessor is then constructed sequentially by interleaving the textual metadata $\{w_i\}$ and the adapted KV-caches using a specialized vocabulary of control tokens (\Cref{fig:contiguous-repr}):
\begin{itemize}
    \item \textbf{Preamble Text}: The sequence begins with a description of the overarching compositional objective. In doing this, we attempt to leverage the language model's text comprehension and instruction following abilities to prime it about the task at hand.
    \item \textbf{Interleaved Bundles}: For each of the $N$ source bundles, the text metadata (Source Text $w_i$) is prepended by a \texttt{<src\_start>} token. This text is immediately followed by its corresponding adapted parametric module $KV_i'$, structurally bounded by \texttt{<kv\_start>}  and \texttt{<kv\_end>} control tokens. These bundle representations are appended sequentially.
    \item \textbf{Combination Text}: Following the $N$ bundles, a \textbf{Combination Text} prompt is appended to provide additional context for the composition. This additional context may include (i) text describing how the source set of tasks relate to the target task, either independently or in concert with each other (ii) examples of the target task placed in context (iii) a text description of the target task.
    \item \textbf{Generation Block}: Finally, the Combination Text is succeeded by a \texttt{<gen\_start>} token to mark the beginning of K-V cache synthesis. A fixed-length sequence of placeholder control tokens ($z_{1}, \dots, z_{L}$) is appended, culminating with a \texttt{<gen\_end>} token.
\end{itemize}

\looseness-1 The entire constructed block is passed to the coprocessor LLM within \tempname for a forward pass. To extract the synthesized parametric skill, we isolate the KV-caches corresponding to the placeholder latent tokens ($z_1, \dots, z_L$), yielding the raw output weights $KV_{out}$. After applying transformations to strip positional information (inverse RoPE de-rotation), $KV_{out}$ is passed through Out K-V Adapter (MLP) to output the synthesized weights $m_{new}$, which can be deployed directly to a frozen downstream base model $M_{\phi}$ to solve $T_{new}$.

\subsection{End-to-End Meta-Training of \tempname}
\label{subsec:training_method_name}

We take an end-to-end approach to train \tempname. We bootstrap the pre-constructed library, $\mathcal{T}_{src}$, to build a meta-training $\mathcal{D}^{\mathrm{train}}$. Each entry in $\mathcal{D}^{\mathrm{train}}$ consists of a structured tuple containing, for each $T \in \mathcal{T}_{src}$: 

\begin{table*}[htbp]
\centering
\renewcommand{\arraystretch}{1.5}
\begin{tabular}{r c p{8.5cm}}
\hline
Component & & \textbf{Description} \\
\hline
$T\{\boldsymbol{x}, \boldsymbol{y}\}$ & & The original task text pairs (inputs $x$ and targets $y$), excluding any pre-existing learned PEFT weights. \\
$\mathcal{T}_{src}\big[T\big]$ & & The ground-truth or retrieved source task mappings for $T$, contributing both their source text ($w_i$) and PEFT modules ($m_i$).  \\
\textbf{Combination Text} ($w$) & & Textual metadata explaining $T$ independently or capturing its relational mapping to the source tasks. \\
\hline
\end{tabular}
\label{tab:task_definitions}
\end{table*}
A concrete example of an entry in $\mathcal{D}^{\mathrm{train}}$ can be found in Appendix \ref{appendix:training_composer_td_eg}. Note that \textbf{Combination Text}($w$) can be the original text meta-data for $T$ (which discusses $T$ independently) or we can generate new text that surfaces information about the relationships between $T$ and $\mathcal{T}_{src}\big[T\big]$.

Having constructed $\mathcal{D}^{\mathrm{train}}$, we learn the parameters of \tempname, denoted by $\theta$, by minimizing the following objective:
\begin{equation*}
    \theta^*= \mathrm{argmin}_{\theta} \sum_{ \big(T\{\boldsymbol{x}, \boldsymbol{y}\}, ~\mathcal{T}_{src}\big[T\big], ~ w \big) ~\sim~ \mathcal{D}^{\mathrm{train}}} \mathcal{L}\left( M_{\phi}(\mathbf{x}; m_{T}), \mathbf{y} \right)
\end{equation*}
where $m_{T} = \mathrm{\tempname}_{\theta}(\{ b_k \}_{k \in \mathcal{T}_{src}[T]}, w)$ is the generated KV-cache. Input to $\tempname(\cdot)$ is constructed as discussed in Section \ref{subsec:_achitecture_}. $\mathcal{L}$ depends on the task and setting, e.g. regular cross-entropy loss or a policy loss for reinforcement learning. For the remainder of this work, we assume that $\mathcal{L}$ is the cross-entropy loss. Note that though we backpropagate through $M_{\phi}$ to get the gradients for $\theta$, we always keep $\phi$ fixed.

\section{Experiments}
To validate the efficacy of \tempname at bridging the text and weight-space modalities, we evaluate our approach across three distinct benchmarks. Below, we outline our experimental workflow, detail our data environments, categorize our baselines by their respective modality limitations, and define our evaluation metrics.

\subsection{Experimental Setup and Workflow}
\subsubsection{Dataset Preparation}
\label{subsubsec:data_prep}
\looseness-1 \noindent We first build our library of source tasks $\mathcal{T}_{src}[T] = \{T_1,  \ldots T_N\}$. For the remainder of this work, we assume $N = 2$. Given the dataset $\mathcal{D}(T_i)= \{\{\mathbf{x}, \mathbf{y}\}, w\}$, we train a prefix K-V cache with $\mathbf{x}$ as the input text and $\mathbf{y}$ as the target over which we compute a negative log-likelihood (NLL) loss \footnote{Note that for all prefix K-Vs we train, we only train for the global layers. Prefix K-Vs with local layers eventually falls out of context}. To inject diversity into our library and discourage overfitting, the input prefix-cache sequence length for each source task is randomly sampled from $\{32, 64, 128\}$. We optimize these modules over a hyperparameter grid of optimization steps ($200, 500, 1000$) and learning rates ($1e^{-2}$ to $1e^{-4}$), selecting the configuration that minimizes validation NLL to form the final task module $m_i$. We instantiate $w_i$, the \textbf{Source Text}, as a description of the task along with few (4-16) in-context demonstrations of the task.

\looseness-1 With  $\mathcal{T}_{src}$ prepared, we construct the training dataset for \tempname following the procedure outlined in \ref{subsec:training_method_name}. If ground-truth task mappings are available (as with Composite-SNI), $\mathcal{T}_{src}[T_{new}]$ is directly mapped. In their absence, we employ the heuristic task retrieval pipeline described in  \ref{subsubsection:task-retrieval} to construct $\mathcal{T}_{src}\big[T_{\mathrm{new}}\big]$. We obtain \textbf{Combination Text} by prompting Gemini 2.5 (see Appendix \ref{appendix:rationale_prompt}) to describe semantic relationships between $\mathcal{T}_{src}[T_{new}]$ and $T_{new}$.

\subsubsection{\tempname Training} 
We train \tempname on the objective described in Section \ref{section:methodology} with cross-entropy loss—delegating the exact hyperparameter choices to Appendix \ref{appendix:training_composer}. To introduce additional regularization, we dynamically sample the output length of the generated target cache from $\{16, 32, 64, 128\}$ during training. For evaluation, we fix the length of the generated prefix-cache to be 32. We use Gemma 3 4B as the LLM for the trainable coprocessor within \tempname. 

\subsubsection{Evaluation Protocol}
For each evaluation task $T_{\mathrm{eval}}$, we construct the target evaluation tuple ($T_{\mathrm{eval}}\{\mathbf{x}, \mathbf{y}\}, \mathcal{T}_{src}\big[T_{\mathrm{eval}}\big]$, \textbf{Combination Text}). All evaluated methods have access to this tuple and may choose to leverage the text fields or weights to generate the solution prefix-cache $m_{T_{\mathrm{eval}}}$. 

\looseness-1 To ensure a fair comparison between \tempname and the baselines, we retrain the prefix weights of the source tasks for each $T_{eval}$ such that they have length 32 and set the output cache size of \tempname and each baseline method\footnote{except for the Concat baseline which ends up with prefix-length 64.} to be 32. Performance is reported on the heldout-set of $T_{\mathrm{eval}}$. We explore evaluating $m_{T_{\mathrm{eval}}}$ in both zero-shot and fine-tuning settings. If fine-tuning, we execute the same grid protocol during Section \ref{subsubsec:data_prep} above, sweeping 4 randomly sampled configurations and keeping the configuration with the best validation performance.

\subsection{Data Environments and Benchmarks}
\subsubsection{Synthetic Task Composition: Composite-SNI (CSNI) Dataset}
\label{composite-sni-dataset}
In order to effectively study mixed-modality composition, we need a robust validation environment where ground-truth task lineages are known. This allows us to systematically isolate and ablate \tempname's synthesis capability without the confounding factor of source task retrieval quality. Using Super-natural Instructions (SNI) \citep{wang2022super} as a foundation, we created a synthetic dataset, Composite-SNI, by presenting pairs of SNI tasks ($T_1, T_2$) to Gemini 2.5 Pro \citep{gemini25}.

Specifically, we prompted it to generate a new composite task, $T_{\{1, 2\}}$, that naturally leverages a subset of the skills needed to solve $T_1$ and $T_2$ independently (see Appendix \ref{appendix:comp_sni_data_construction}). Our final Composite-SNI dataset consists of $\approx21$K composite tasks, which we split into meta-train ($\approx 17$K tasks seen by \tempname during the training phase) and meta-eval sets. To evaluate the ability of \tempname to generalize beyond the $\mathcal{T}_{src}$ encountered during training (c.f. \Cref{sec:method_generalization}), we further stratify the meta-eval set into three subsets based on whether their constituent source tasks appear in the meta-train set: (i) \textbf{Both-Seen}: Both source tasks appear in at least one meta-train composite task; (ii) \textbf{One-Seen}: Only one source task appears in the meta-train set; and (iii) \textbf{Neither-Seen}: Neither source task appears in the meta-train set.

\subsubsection{\tempname in the Wild: Datasets and Heuristic Source Task Retrieval}
\label{subsubsection:task-retrieval}
In real-world deployment scenarios, the ground-truth source tasks required to synthesize a novel target capability are rarely known. Thus, while Composite-SNI is an effective foundation for understanding \tempname, we need to investigate our mix-modality composition approach in realistic scenarios by experimenting with the datasets below.

\noindent \textbf{Super-natural Instructions (SNI) \citep{wang2022super}}: A benchmark of 1,616 diverse natural language processing tasks across 76 semantic categories, i.e. text classification, information extraction, sequence tagging, text writing and composition. We use a subset of 875 tasks, each containing 1024 instances for our experiments. 

\noindent \textbf{MMLU-ProX \citep{mmluprox}:} To stress-test cross-lingual and cross-domain generalization, we leverage this multilingual dataset spanning 14 subject categories and 29 languages across five geographic regions. To construct a challenging evaluation split, we identified the lowest-performing languages for Gemma 3 4B within each region (per Table 6 in the Appendix of \citep{mmluprox}) and held out the three lowest-performing languages overall. Additionally, we randomly excluded three subject categories from training. Hence, the meta-training set comprises 26 languages and 11 categories, while the evaluation pool retains all 29 languages and 14 categories. This yields a two-pronged held-out breakdown: (i) 3 completely unseen languages evaluated across all 14 categories, and (ii) 26 seen languages evaluated across the 3 unseen categories. We evaluate on a final held-out space of 6 randomly sampled language-category pairs: \texttt{wolof\_math}, \texttt{wolof\_health}, \texttt{zulu\_physics}, \texttt{spanish\_law}, \texttt{indonesian\_law}, and \texttt{afrikaans\_history}.

Since the above datasets do not have ground truth source-task mappings, 
we introduce a two-stage heuristic task retrieval pipeline to construct the target to source-task-set mapping:
\begin{enumerate}
    \item \textbf{Semantic Retrieval:} Given a target task $T_{\mathrm{new}}$ and a library of available source tasks $\mathcal{T}_{src}$, we employ a retriever built on Gemini Embeddings \citep{gemini-embedding} to rank the source tasks by their semantic relevance to $T_{\mathrm{new}}$ (see Appendix \ref{appendix:retrieval-details} for architecture and training details). From this ranking, we construct $P$ candidate pairs by grouping the top-ranked tasks sequentially (e.g., the top two form the first candidate pair, the next two form the second, etc.).
    \item \textbf{LLM Selection:} We present the formatted candidate task metadata alongside the target task profile to Gemini 2.5 Pro, which performs a discrete selection of the single most contextually relevant pair (Appendix \ref{appendix:llm-selection}). We take this final pair as $\mathcal{T}_{src}\big[T_{new}\big]$.
\end{enumerate}

\subsection{Baselines}
\label{subsec:baselines}
Given a target task $T_{new}$ paired with source tasks $T_1$ and $T_2$ and their corresponding task bundles $b_{T_1} = (w_{T_1}, m_{T_1})$ and $b_{T_2} = (w_{T_2}, m_{T_2})$, we compare \tempname against standard approaches across the following uni-modal baselines. Note that for all these methods (except ICL), we investigate both zero-shot and continued training on $T_{new}$-only data.

\subsubsection{Weight-Space Only}
These baselines represent standard parameter-space model merging paradigms that combine learned behaviors efficiently but lack the capacity to process semantic instructions or relational task metadata. 
\begin{itemize}
    \item \textbf{LERP (Linear Interpolation):} Performs simple parameter-space task arithmetic by averaging the weights of the two parent prefix caches element-wise:
    \begin{equation*}
        m_{T_{\mathrm{new}}} = \frac{m_{T_1} + m_{T_2}}{2}
    \end{equation*}
    This is conceptually equivalent to task arithmetic \citep{task-arithmetic}.
    \item \textbf{Concat:} Directly concatenates both parent prefix caches along their sequence dimension, yielding a target prefix sequence length of 64 tokens. This method results in prefix-caches that have strictly more capacity than all other methods.
    \begin{equation*}
        m_{T_{\mathrm{new}}} = m_{T_1} \circ m_{T_2}
    \end{equation*}
    \item \textbf{Source-Task Transfer:} Computes the average zero-shot performance of the parent KV-caches on the target task. We also evaluate a prefix-tuning approach that initializes the target KV-cache using the parent KV-caches prior to continuation training. We report the averaged performance of this method, which is functionally equivalent to SPoT \citep{spot}. 
    \item \textbf{Singular Value Decomposition (SVD):} Inspired by \cite{svd-lora}, we  perform SVD across the sequence (\texttt{SVD\_SEQ\_AXIS}), head (\texttt{SVD\_HEAD\_AXIS}), or embedding (\texttt{SVD\_EMBED\_AXIS}) dimensions of the prefix caches. We stack $N$ trained KV-caches along a new terminal dimension and permute the tensor to isolate the target axis $d$ and the checkpoint axis $N$ as the final two dimensions. SVD is applied independently to each $d \times N$ matrix slice ($M = U \Sigma V^T$). To prevent feature collapse and preserve the energy distribution of the parameters, we scale the first left singular vector ($U_{:, 1}$) by the root mean square of all singular values:
    \begin{equation*}
        m_{T_{\mathrm{new}}} = U_{:, 1} \sqrt{\frac{1}{N} \sum_{i=1}^{N} \sigma_i^2}
    \end{equation*}
    The output is subsequently permuted and squeezed back to the original tensor dimensions.
\end{itemize}

\subsubsection{Text-Space Only}
This approach leverages the base language model's default text comprehension capabilities without adapting its parameter space.
\begin{itemize}
    \item \textbf{In-Context Learning (ICL):} Evaluates the performance of $k$-shot ICL, where $k$ is selected from $\{16, 32, 64\}$ based on validation set NLL. For tasks with constrained sample numbers, $k$ is capped at the closest power of 2.
\end{itemize}

\subsubsection{Transfer-less Adaptation}
\begin{itemize}
    \item \textbf{Direct Prefix Tuning:} Optimizes a fresh target task prefix-cache from scratch on $T_{new}$ under standard initialization, or initialized using the first 32 tokens of ICL examples from the target task. As detailed in \Cref{sec:meta-data-ablation}, we also compare against a text-augmented direct baseline where all textual metadata available to \tempname (source texts and combo-text) is directly prepended to the input examples before prefix-tuning.
\end{itemize}

\subsection{Evaluation Metrics: Global Elo and Absolute Scale Verification}
For each task, we measure the model's average negative log-likelihood (NLL) loss of the targets given its inputs on the heldout set. Because evaluation benchmarks vary significantly in scale and numerical boundaries, aggregating raw NLL values directly can bias results toward high-variance tasks. To provide a scale-invariant and balanced global evaluation, we map model performance into an adapted global Elo rating system where each method is treated as an individual competitor in a tournament pool, initialized at $R = 1500.0$.  We compute Elo scores via a two-step optimization process:
\begin{enumerate}
    \item \textbf{Empirical Win-Rate Matrix Estimation:} We construct an $N \times N$ empirical win-rate matrix $W$, where $N$ represents the total number of competitive methods. For any pair of distinct methods $i$ and $j$, the observed win rate $W_{ij}$ represents the empirical fraction of evaluation tasks where method $i$ strictly outperforms method $j$ (achieving a lower NLL score):
    \begin{equation*}
        W_{ij} = \frac{1}{| \mathcal{T}_{eval} |} \sum_{T \in \mathcal{T}_{eval}} \mathbb{I}(\mathcal{L}_i(T) < \mathcal{L}_j(T))
    \end{equation*}
    where $\mathcal{L}_i(T)$ represents the evaluation loss of method $i$ on task $T$, and $\mathbb{I}(\cdot)$ denotes the indicator function. Self-comparisons ($W_{ii}$) are uniformly set to $0.5$.
    \item \textbf{Global Rating Optimisation:} Under the Bradley-Terry framework, the latent expected win rate $E_{ij}$ of method $i$ over method $j$ is modeled as a logistic function of their underlying latent Elo ratings $R_i$ and $R_j$:
    \begin{equation*}
        E_{ij} = \frac{1}{1 + 10^{(R_j - R_i)/400}}
    \end{equation*}
    To extract the optimal latent scores that minimize variance across empirical interactions, we formulate rating estimation as a non-linear objective function minimizing the sum of squared errors between the observed and expected win rates:
    \begin{equation*}
        \min_{\mathbf{R}} \sum_{i=1}^{N} \sum_{j \neq i} \left( W_{ij} - \frac{1}{1 + 10^{(R_j - R_i)/400}} \right)^2
    \end{equation*}
    This objective function is optimized numerically using the L-BFGS-B algorithm \citep{l-bfgs-b}, where the $i$-th entry of the optimized vector $\mathbf{R}$ defines the final global Elo rating for the $i$-th method.
\end{enumerate}

Along with the reported Elo Scores in Section \ref{sec:results_and_disc}, we present the unaggregated raw NLL scores for every baseline and task in Appendix \ref{appendix:raw_nll_scores} for transparency and verification alongside the relative rankings.
\section{Results}
\label{sec:results_and_disc}
\subsection{\tempname learns to synthesize text and weight-space artifacts when ground-truth task relationships are known.}
\begin{figure}[h!]
\begin{center}
 \includegraphics[scale=0.4]{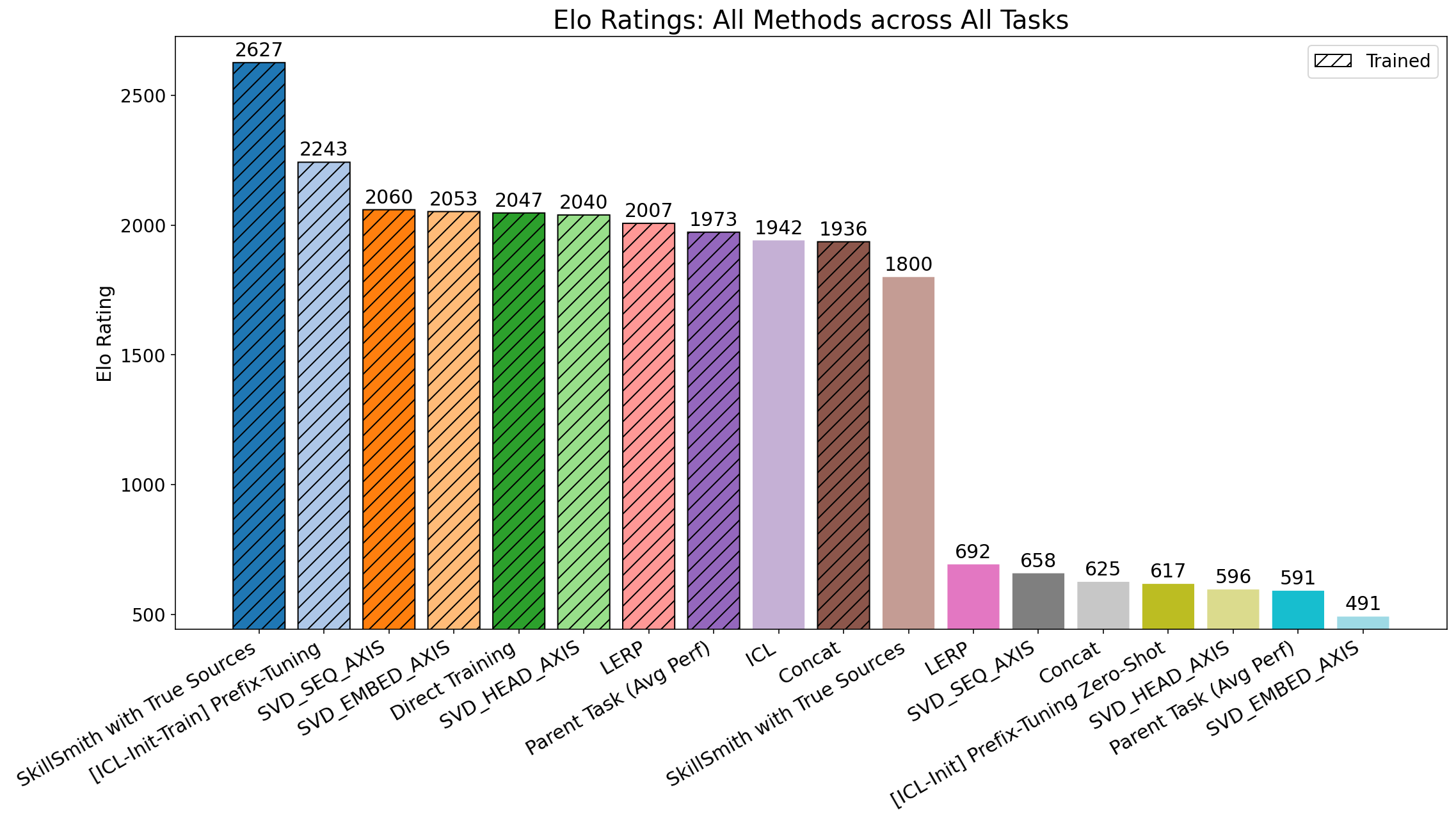}
\caption{\looseness=-1 ELO ratings of various methods, higher is better. Across 15 meta-test tasks from the Composite-SNI dataset, \tempname outperforms a slew of baselines.}
\vspace{-5mm}
\label{fig:baseline_compsni}  
\end{center}
\end{figure}

In this section, we analyse the performance of \tempname in the setting of the Composite-SNI, where ground truth source tasks are known. This allows us to perform an analysis without being confounded by noise when selecting the source task set via approximate retrieval. 


\Cref{fig:baseline_compsni} demonstrates the effectiveness of \tempname in synthesizing text and prefix-cache information from ground-truth source tasks to solve novel target capabilities. In the zero-shot regime, where we evaluate the synthesized prefix-caches before any direct training on the downstream task, \tempname solidly outperforms all baseline weight-space merging techniques. We do note that standard in-context learning (ICL) remains highly competitive in the zero-shot regime, indicating that the base model maintains an edge over zero-shot parametric generation without any adaptation. 

\looseness-1 When we allow for task-specific fine-tuning, however, \tempname establishes as a clear winner across all evaluated methods. Fine-tuning the prefix-weights initialised by \tempname yields representations that drastically outperform both the strongest non-compositional baseline (prefix-tuning initialized via ICL) and continuing training after traditional arithmetic weight-merging. These results underscore that natively reasoning over both textual metadata and parametric skills provides superior parameter initialization, allowing the model to adapt to target tasks much more effectively than relying on either modality in isolation.

\subsection{\tempname is effective in the wild}
Composite-SNI served as an ideal test bed for \tempname because it has a relatively large number of meta-train tasks ($\sim17$K) and the ground truth set of source (from SNI) tasks are known.  Using the SNI and MMLU-ProX datasets, we investigate \tempname's ability to handle settings where (i) we are meta-task poor (SNI has $\sim 800$ tasks whilst MMLU-ProX has $\sim 250$) and (ii) the relationship between tasks can only be inferred since there are no ground truth source task mappings. 

We rely on the retrieval based approach outlined in Section \ref{subsubsection:task-retrieval} to map each task to a distinct pair of parent source tasks. For any meta-training task $T \in \mathcal{T}_{train}$, its source tasks must be drawn from training pool excluding itself: $\mathcal{T}_{src} \subset \mathcal{T}_{train} \setminus \{ T \}$. Further, for any meta-evaluation task $T \in \mathcal{T}_{eval}$, its source tasks must also be drawn from the training set: $\mathcal{T}_{src}[T] \subset \mathcal{T}_{train}$ with $\mathcal{T}_{train} \cap \mathcal{T}_{eval} = \emptyset$. This simulates a realistic setting where the agent can improve offline on what it has seen and use its experiences in the past to solve a current task. To address training \tempname in the meta-task poor regimes of MMLU-ProX and SNI, we also investigate bootstrapping the \tempname trained for the Composite SNI task by using it as initialization. 

\subsubsection{Super-natural Instructions}

\begin{figure}[h!]
\begin{center}
 \includegraphics[width=\linewidth, height=9cm, keepaspectratio]{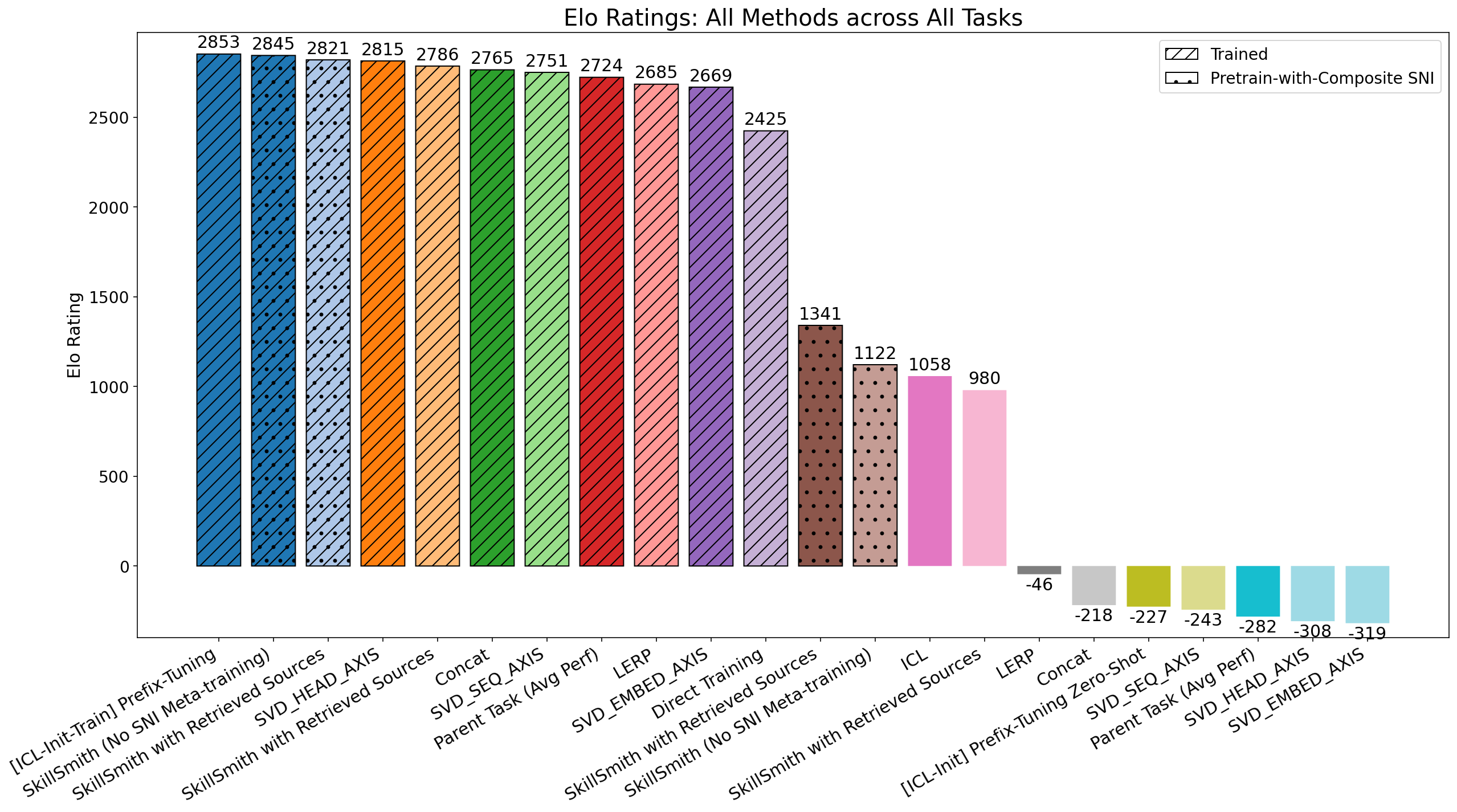}
\caption{Elo scores are computed over 10 SNI tasks used as meta-test tasks. These tasks were strictly heldout and also did feature in the construction of CSNI. Retrieving a noisy set of parent tasks, combining their prefix caches and text meta-data via \tempname is competitive.}
\label{fig:sni_results}  
\vspace{-5mm}
\end{center}
\end{figure}

\Cref{fig:sni_results} demonstrates \tempname's efficacy even under the constraints discussed above. In the zero-shot regime, the trends mirror those of \Cref{fig:baseline_compsni}: \tempname variants and the ICL baseline outperform all other methods. Training \tempname exclusively on SNI data without bootstrapping from a pre-trained Composite-SNI checkpoint underperforms ICL. However, evaluating the raw CSNI checkpoint directly on the original SNI tasks (\tempname (No SNI Meta-Training)) outperforms ICL, showing that the composition capabilities acquired on synthetic CSNI data natively transfer to an organic task distribution.  The best-performing zero-shot method is achieved by fine-tuning the CSNI checkpoint with SNI data, which addresses the concern of data scarcity whilst performing in-domain adaptation to the SNI data distribution.

In contrast with the zero-shot results, task-specific fine-tuning on downstream data yields marginal separation in Elo ratings among the top three evaluated methods. The win rate between these methods converges to 0.5, indicating no significant performance differences. We posit that this performance convergence is likely attributable to two factors: first, SNI tasks feature significantly higher instance counts per task ($O(1000)$) relative to Composite-SNI; second, because SNI was established in 2022 (\citep{wang2022super}), it primarily comprises primitive NLP tasks—such as simple sentiment classification or character concatenation—that are arguably trivial for contemporary base language models. These two effects results in the top fine-tuning based methods -- including \tempname -- to cluster around the same performance ceiling.

\subsubsection{MMLU-ProX}
\begin{figure}[t!]
\begin{center}
 \includegraphics[width=\linewidth, height=9cm, keepaspectratio]{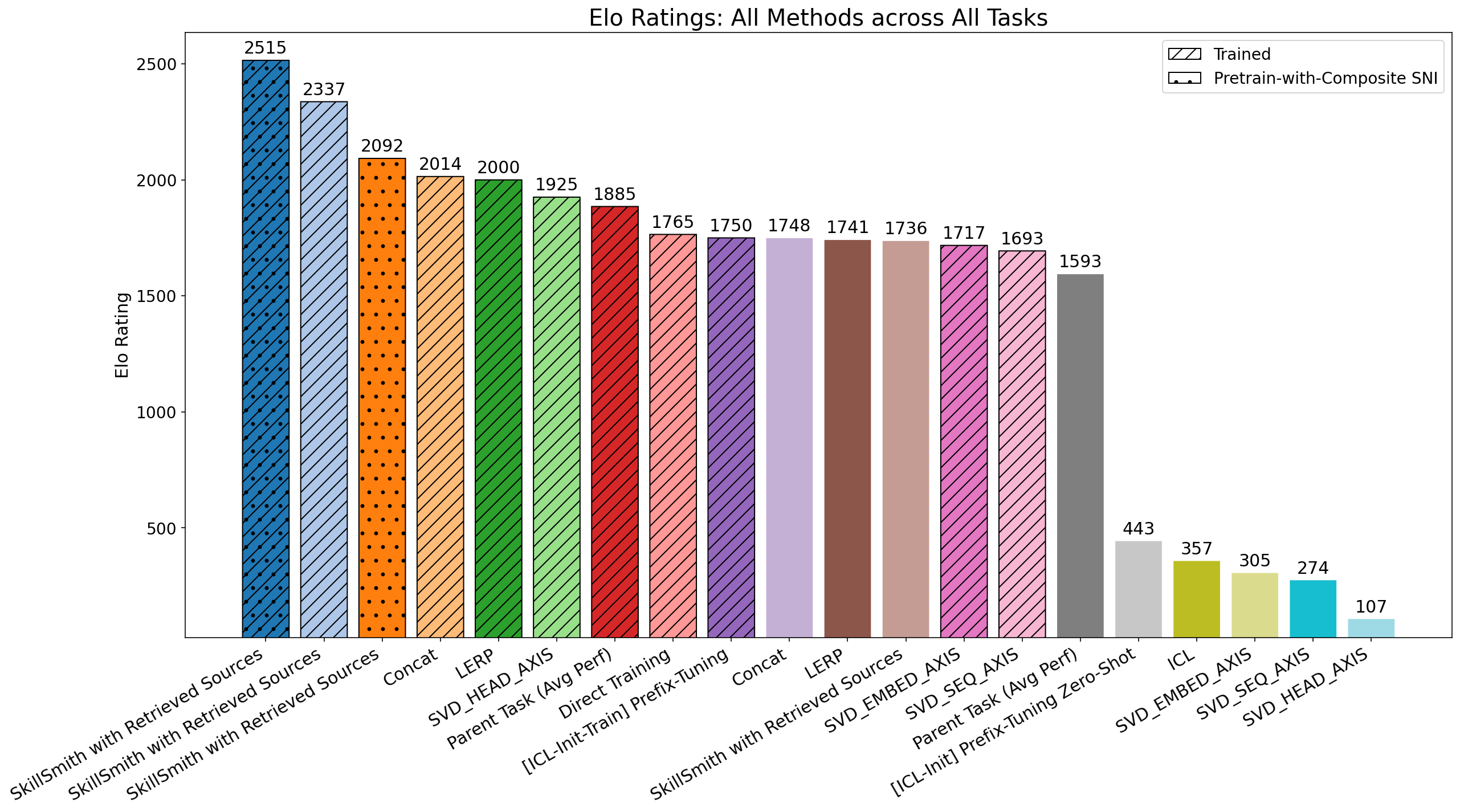}
\caption{Elo scores are computed over 6 MMLU-ProX tasks used as meta-test tasks. }
\label{fig:mmlu_results}  
\vspace{-6mm}
\end{center}
\end{figure}

\looseness-1 Unlike SNI, MMLU-ProX has no genealogical relationship with the CSNI, yet consistent with the SNI results, zero-shot [Pretrain-CSNI] \tempname with Retrieved Sources, which uses the CSNI checkpoint as an initialisation to be trained on MMLU source tasks, outperforms all zero-shot baselines. More surprisingly, this configuration also outperforms all baselines that are allowed to \textit{fine-tune} on downstream tasks. We can see concretely the value of \textit{pre-training} \tempname on synthetic compositions, as training \tempname exclusively on the limited MMLU-ProX tasks without bootstrapping (second column from the left) yields a substantially lower zero-shot Elo of 1736.

\looseness-1 While the downstream fine-tuning results on SNI plateaued due to task simplicity, the MMLU-ProX dataset reveals an evaluation regime where fine-tuning widens the performance gap. Fine-tuning the bootstrapped SkillSmith trained on MMLU task sources achieves the highest overall ELO rating, maintaining a clear margin over both the unbootstrapped model variant and the strongest weight-space baseline (Concat).

\looseness-1 We attribute that this pronounced performance gap stems from the dataset's sparsity and inherent difficulty. MMLU-ProX contains only roughly 250 tasks total, and our evaluation targets subject categories across low-performing languages for the fixed, downstream model. Because these tasks are harder than SNI and lack an abundance of task instances, the methods do not converge around a performance ceiling. Under data limits, standard gradient descent cannot easily compensate for suboptimal initialisations. The superior parameter initialisation synthesised by the bootstrapped \tempname provides an advantage for downstream adaptation that direct training or uniform weight merging cannot match.

\section{Analysis}
\subsection{Does \tempname actually use the K-V caches?}
\begin{table}[htbp]
\centering
\caption{ELO ratings between different versions of \tempname trained with varying input content evaluated on the Composite-SNI meta-eval tasks.}
\label{tab:input_ablation}
\begin{tabular}{lc}
\toprule
\textbf{Input Configuration} & \textbf{Elo Rating} \\
\midrule
No inputs & 1209 \\
Only K-V Caches & 1455 \\
All Inputs except K-V Caches & 1622 \\
\textbf{All Inputs} & \textbf{1714} \\
\hline\hline
\end{tabular}
\vspace{-5mm}
\end{table}
\looseness-1 While we provide source task prefix caches as input to \tempname, it is possible that \tempname does not actually leverage this information. This would call into question the our central objective of learning to reason over both text and parametric weights. We therefore ablate the differential impact on performance of dropping out different parts of the input to \tempname.

\Cref{tab:input_ablation} shows that \tempname learns to actually use both text meta-data and parametric information to generate prefix caches. Using only text-metadata outperforms using only weight space inputs, but results are best when both input types are present. Note that the `Only K-V Caches` entry mimics ATTEMPT \citep{attempt} since it corresponds to learning to combine PEFT modules using a parametric function instead of simple arithmetic operations like averaging and concatenation.

\subsection{Where do \tempname's performance gains come from?}
\label{sec:meta-data-ablation}
A critical question is whether \tempname's superior performance (as seen in \Cref{fig:baseline_compsni} and \Cref{fig:mmlu_results}) stems from its architectural ability to synthesise modalities, or simply from having access to richer textual context—an advantage that weight-only baselines like LERP and Concat lack.

\begin{wrapfigure}[19]{l}{0.455\textwidth}
\vspace{-4mm}
    \centering
    \includegraphics[scale=0.28]{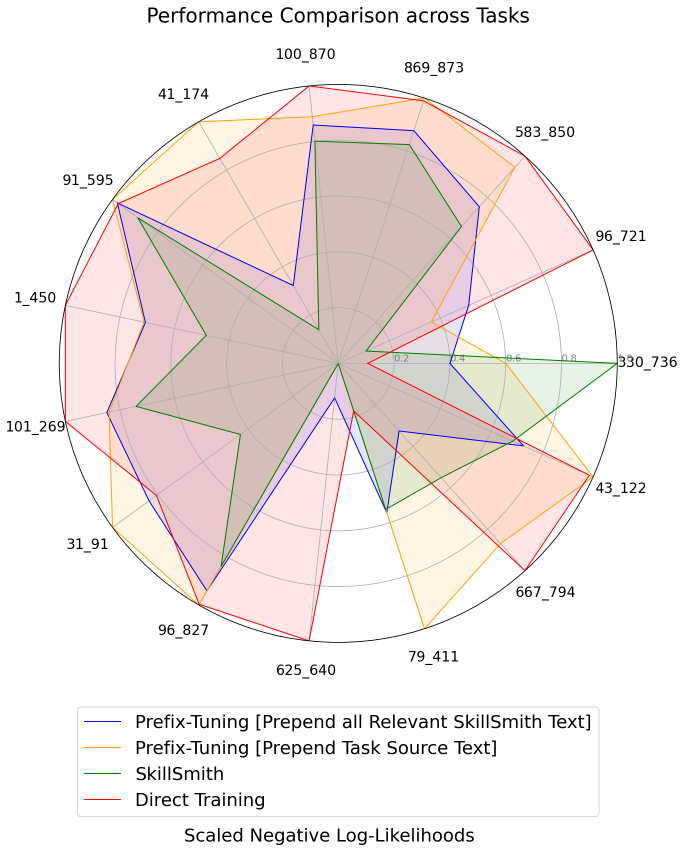}
    \caption{Smaller covered area is better. NLLs scaled to (0, 1) for clarity.}
    \label{fig:text-metadata-ablation}
\end{wrapfigure}

\looseness=-1 To isolate this variable, we conducted an ablation study across our held-out Composite-SNI target tasks. We extracted all the textual information provided to \tempname (specifically, the source texts $w_{i}$ and the combination-text) and directly prepended it to the target task's input examples. We then trained a standard prefix-cache on this text-augmented dataset.

\looseness=-1 As illustrated by the spider plot in \Cref{fig:text-metadata-ablation}, while providing auxiliary text generally improves the Direct Training baseline, it consistently falls short of matching \tempname's performance across the task set. This performance gap demonstrates that \tempname's gains are not solely attributable to the mere presence of textual metadata. Rather, \tempname is successfully learning a synergistic composition of the textual instructions and the parametric weights, allowing it to generate more effective task representations than text-augmented direct training alone.

\subsection{Does \tempname actually generalize?}
\label{sec:method_generalization}
\begin{figure}[h!]
\begin{center}
 \includegraphics[scale=0.275]{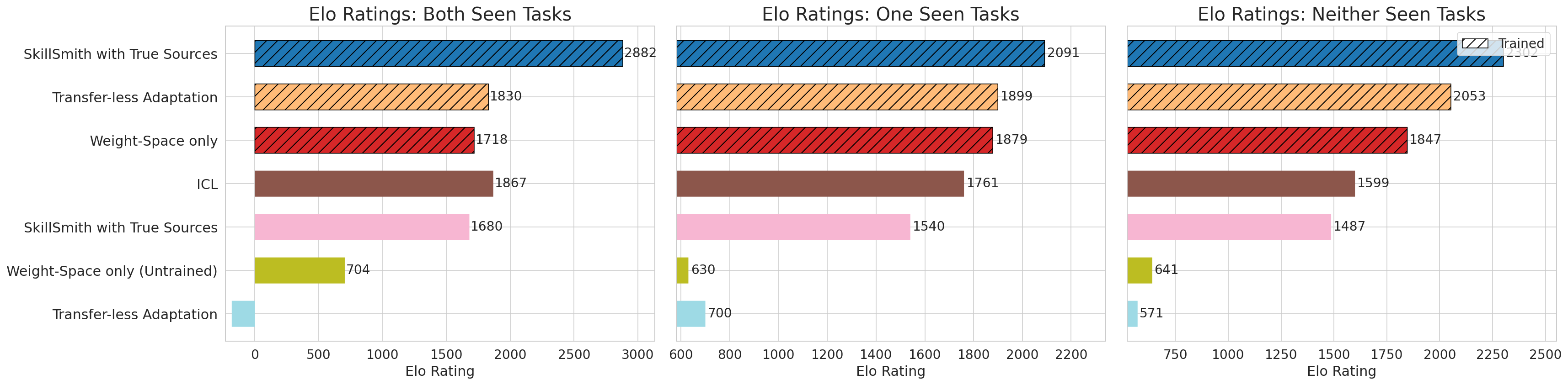}
\caption{\looseness=-1 We further break down the 15 CSNI meta-test tasks into the 3 categories defined \Cref{composite-sni-dataset}. We also aggregate methods according the groups in \Cref{subsec:baselines}.}
\vspace{-5mm}
\label{fig:comp_sni_generalization}  
\end{center}
\end{figure}

The results of \Cref{fig:baseline_compsni} are aggregated over 15 tasks, 5 each from the original CSNI sub-splits (\textbf{Neither-seen}, \textbf{One-seen}, \textbf{Both-seen}). To ensure if \tempname actually generalizes beyond the tasks seen during meta-training, we re-compute the evaluation per sub-split.

Across all splits, \tempname is the clear winner. Even when faced with composite tasks where none of their parent tasks were used to construct the tasks in the meta-train set,  \tempname achieves the best Elo rating by a large margin. This attests to \tempname's ability to generalize beyond its training distribution.

\section{Conclusion}
We believe that as agents become more pervasive and capable, they need to be equipped with the appropriate set of tools to attack increasingly complex tasks. This work has introduced one such tool: \tempname, an augmented LLM architecture designed to bridge the gap between textual reasoning and parametric adaptation. By treating weight-space inputs as a native modality, \tempname facilitates the direct synthesis of task-specific prefix-weights steered by high-level textual instructions and metadata.  

\looseness-1 Our evaluation on the Composite-SNI dataset demonstrates that \tempname\ consistently outperforms standard weight-space merging techniques, such as LERP and Concat, in zero-shot settings. Furthermore, it provides a significantly stronger starting point for fine-tuning compared to random or ICL-based initialization methods. In this regime, \tempname shows significant promise in capturing the complex functional relationships across modalities that simple arithmetic baselines cannot.  

\looseness-1 Moving beyond idealized conditions, our experiments on the standard SNI and MMLU-ProX datasets highlight \tempname's effectiveness in "in-the-wild" scenarios where ground-truth task mappings are unavailable. While our findings on SNI indicate that abundant downstream training data on simpler tasks allows standard baselines to reach a similar performance ceiling, \tempname proves remarkably advantageous in challenging, data-sparse environments. On the MMLU-ProX benchmark, the superior parameter initialization synthesized by a bootstrapped \tempname provides an advantage for downstream adaptation that direct training or weight merging cannot match. This demonstrates that \tempname transfers composition capabilities acquired on synthetic data to organic task distributions, effectively overcoming data limits and the potential adverse effects of noisy retrieval. 

Ultimately, \tempname represents a step toward more holistic agentic architectures—systems that no longer view "knowing" (text) and "doing" (weights) as orthogonal, but as a unified basis for instruction-steered adaptation. 

\clearpage

\bibliography{main}
\newpage
\appendix
\section{Composite SNI Dataset Construction}
\label{appendix:comp_sni_data_construction}
We construct Composite-SNI a synthetic dataset, where each task $t_1\_t_2$ is constructed by combining pairs of tasks $t_1$ and $t_2$ from the SNI \citep{wang2022super} dataset. Below, we outline the steps we took to construct this dataset and to ensure that the final set of generated tasks are of high quality.
\subsection{Data Generation Procedure}
\subsubsection{Initial Task Generation}
We first create a set of all pairs of tasks from the base SNI dataset. We prompt Gemini 2.5 Pro with each of pair of tasks (task description and few-shot examples) and ask for it to generate a new task that leverages the underlying set of skills required to solve the individual tasks in the original pair. We also prompt the model to give a rationale for why the generated task is a good one, and a score from 1-5 of the perceived quality of the generation.  This step results in O(350K) generated tasks. Below is the full prompt that we use.

\begin{promptbox}[prompt:synthetic]{Synthetic Dataset Generation Task}

You are a polymath, tasked with creating a synthetic instruction-following dataset. Your expertise spans a wide range of domains and technical areas, allowing you to find deep connections between disparate tasks.

Your specific task is to take a set of existing instructions and their example solutions, and synthesize them into a new, well-thought-out instruction. This new instruction must be an interesting, realistic, and relatively challenging task that a language model could reasonably solve. It should represent a deep, semantic, and logical combination of the provided seed instructions.

For the new instruction you create, YOU MUST ALSO PROVIDE 128 TO 512 EXAMPLE SOLUTIONS.

After creating the combined instruction, you will score its quality on a scale of 1 (very poor) to 5 (excellent) and provide a rationale for your score.

Your input will follow this exact format:

\vspace{1em}
\noindent \textbf{\# Begin Input Instructions \#}

\vspace{0.5em}
\noindent \textbf{\#\# Instruction 1 \#\#} 

\noindent \textbf{\#\#\# Begin Task Description \#\#\#} 

\noindent ((task description goes here)) 

\noindent \textbf{\#\#\# Example Begin \#\#\#} 

\noindent \textbackslash{}t[Input]: ((example goes here)) 

\noindent [Targets]: ((example goes here)) 

\noindent \dots 

\noindent \textbf{\#\#\# Example Begin \#\#\#} 

\noindent \textbackslash{}t[Input]: ((example goes here)) 

\noindent [Targets]: ((example goes here))

\vspace{0.5em}
\noindent \textbf{\#\# Instruction 2 \#\#} 

\noindent \textbf{\#\#\# Begin Task Description \#\#\#} 

\noindent ((task description goes here)) 

\noindent \textbf{\#\#\# Example Begin \#\#\#} 

\noindent \textbackslash{}t[Input]: ((example goes here)) 

\noindent [Targets]: ((example goes here)) 

\noindent \dots 

\noindent \textbf{\#\#\# Example Begin \#\#\#} 

\noindent \textbackslash{}t[Input]: ((example goes here)) 

\noindent [Targets]: ((example goes here))

\vspace{1em}
\noindent \textbf{\# End Input Instructions \#}

\vspace{1em}
\noindent Your output should follow the following format:

\vspace{1em}
\noindent \textbf{\# Begin Output \#}

\vspace{0.5em}
\noindent \textbf{\#\# Begin Instruction \#\#} 

\noindent ((task description goes here)) 

\noindent \textbf{\#\#\# Example Begin \#\#\#} 

\noindent \textbackslash{}t[Input]: ((example goes here)) 

\noindent \textbackslash{}t[Targets]: ((example goes here)) 

\noindent \dots 

\noindent \textbf{\#\#\# Example Begin \#\#\#} 

\noindent \textbackslash{}t[Input]: ((example goes here)) 

\noindent \textbackslash{}t[Targets]: ((example goes here)) 

\noindent \dots

\vspace{0.5em}
\noindent \textbf{\#\# End Instruction \#\#} 

\noindent \textbackslash{}t[[Score]]: ((score)) 

\noindent \textbackslash{}t[[Rationale]]: ((rationale))

\vspace{1em}
\noindent \textbf{\# End Output \#}

\vspace{1em}
\noindent Note that ellipses means there could be more examples / instructions your can generate. 

\noindent PLEASE STICK TO THIS OUTPUT FORMAT EXACTLY ELSE WE WILL HAVE DOWNSTREAM PARSING ERRORS ! 

\noindent DO NOT FORGET YOU MUST ALSO PROVIDE 128 TO 512 EXAMPLE SOLUTIONS. 

\noindent If an example of the instruction does not have a target, you can leave it blank 

\noindent Here is an example to guide you:

\vspace{1em}
\noindent \textgreater\textgreater\textgreater{} User Input: 

\noindent \textbf{\# Begin Input Instructions \#}

\vspace{0.5em}
\noindent \textbf{\#\# Instruction 1 \#\#} 

\noindent \textbf{\#\#\# Begin Task Description \#\#\#} 

\noindent Give an example of an object that is spherical in shape 

\noindent \textbf{\#\#\# Example Begin \#\#\#} 

\noindent \textbackslash{}t[Input]: basket ball 

\noindent [Targets]: 

\noindent \textbf{\#\#\# Example Begin \#\#\#} 

\noindent \textbackslash{}t[Input]: orange 

\noindent [Targets]: 

\noindent \textbf{\#\#\# Example Begin \#\#\#} 

\noindent \textbackslash{}t[Input]: eyeball 

\noindent [Targets]:

\vspace{0.5em}
\noindent \textbf{\#\# Instruction 2 \#\#} 

\noindent \textbf{\#\#\# Begin Task Description \#\#\#} 

\noindent Write a short poem a comedian 

\noindent \textbf{\#\#\# Example Begin \#\#\#} 

\noindent \textbackslash{}t[Input]: There was once a funny man, who did funny things, he slipped and fell. 

\noindent [Targets]: The whole crowd laughed

\vspace{1em}
\noindent \textbf{\# End Input Instructions \#}

\vspace{1em}
\noindent \textgreater\textgreater\textgreater{} Your Output: 

\noindent \textbf{\# Begin Output \#}

\vspace{0.5em}
\noindent \textbf{\#\# Begin Instruction \#\#} 

\noindent Present a joke a comedian would tell about a ball 

\noindent \textbf{\#\#\# Example Begin \#\#\#} 

\noindent \textbackslash{}t[Input]: I tried to start a juggling club, but I couldn't get the ball rolling. After many failed attempts, I decided to just throw in the towel. 

\noindent \textbackslash{}t[Targets]:

\vspace{0.5em}
\noindent \textbf{\#\# End Instruction \#\#} 

\noindent \textbackslash{}t[[Score]]: 5 

\noindent \textbackslash{}t[[Rationale]]: This is a good instruction because it recognizes that a poem by a comedian can be a joke, and also realises that a ball has spherical shape. The example given is also a good one that semantically integrates the essence of the two instructions.

\vspace{1em}
\noindent \textbf{\# End Output \#}

\vspace{1em}
\noindent OK, NOW HERE WE GO WITH AN ACTUAL TASK.

\vspace{1em}
\noindent \{\{text\}\}

\end{promptbox}

\subsubsection{Task Filtration}
Given the large number of tasks generated, we take the following steps to prune the list down to only a highest quality set of generations:
\begin{enumerate}
    \item We remove all tasks that have a self-reported (as part of the task generation) quality score $\leq 3$.
    \item We remove all tasks that have < 16 accompanying generated task examples.
    \item We perform 2 rounds of LLM-mediated task filtration. Specifically, we group the remaining tasks randomly into pairs, and again prompt Gemini 2.5 to select which of the tasks is better based on a pre-defined rubric with axes such as clarity, actionability and diversity of task examples. We expose our exact prompt below.
\end{enumerate}
These filtration steps result in a final dataset size of O(90K) tasks.
\begin{promptbox}[prompt:evaluator]{LLM Prompt Evaluator Task}

You are an expert evaluator of instruction-following prompts for large language models. Your primary goal is to identify instructions that are strong candidates for inclusion in a dataset focused on interesting and complex, multi-step composite instructions. You will compare two instruction-following examples, each consisting of an instruction (prompt) and examples of how an LLM followed that instruction.

\vspace{0.5em}
\noindent \textbf{Criteria for a Superior Instruction:}

When evaluating, prioritize the following properties in the instruction itself and the quality of its examples:

\noindent \textbf{Clarity and Specificity:} Is the instruction unambiguous, easy to understand, and precise? Does it clearly define the task, expected output format, and any critical constraints?

\noindent \textbf{Conciseness:} Is the instruction brief and to the point, avoiding unnecessary jargon, repetition, or verbose explanations, without sacrificing clarity or specificity?

\noindent \textbf{Actionability:} Does the instruction describe a task that an LLM can perform effectively and directly, leading to a clear, well-defined, and consistent output?

\noindent \textbf{Output Diversity \& Quality:} Does it encourage a robust and diverse range of high-quality, appropriate outputs from the LLM, rather than being overly narrow, trivial, or leading to repetitive responses?

\noindent \textbf{Quality of Examples:} Do the provided examples demonstrate clear, correct, and varied adherence to the instruction? Do they effectively showcase the intended behavior and the range of acceptable outputs?

\vspace{0.5em}
\noindent \textbf{Your Task:}

Evaluate which of the two provided instruction-following examples is better overall, considering all the desirable properties listed above, with a strong emphasis on correctness, diversity and robustness. 

\noindent The instructions will come in the following input template:

\vspace{1em}
\noindent \textbf{\# Begin Input Instructions \#}

\noindent \textbf{\#\# Instruction 1 \#\#} 

\noindent \dots

\noindent \textbf{\#\# Instruction 2 \#\#} 

\noindent \dots

\noindent \textbf{\# End Input Instructions \#}

\vspace{1em}
\noindent Where the ellipses (\dots) represent the actual instruction-following task.

\vspace{0.5em}
\noindent \textbf{Output Format:}

\noindent [[Rationale]]: ((reason)) 

\noindent [[Result]]: ((output))

\vspace{0.5em}
\noindent Return 0 as output if the first instruction is superior. 

\noindent Return 1 as output if the second instruction is superior. 

\noindent Populate the rationale field with the reasons why you think your chosen instruction is superior to the other. 

\noindent Make sure to keep your rationale short and succint.

\vspace{1em}
\noindent PLEASE STICK TO THIS OUTPUT FORMAT EXACTLY ELSE WE WILL HAVE DOWNSTREAM PARSING ERRORS ! 

\noindent Strictly adhere to the output format (either '1' or '0').

\vspace{1em}
\noindent Now below is a task for your to solve:

\vspace{1em}
\noindent \{\{text\}\}

\end{promptbox}

\subsubsection{Task Correctness Filtration}
The previous filtration step was based on comparison with other tasks. For this stage, we filtered based on task correctness. We first performed a manual inspection of a random sample of the generated tasks to come up with good examples of violations that render a task incorrect. This manual inspection also allows us to estimate what we should a-priori expect to be a good retention rate after bulk LLM based filtration based on correctness. Informed by the manual inspection of tasks, we construct the following prompt for Gemini 2.5 to select tasks based on correctness, internal consistency and factuality.
\begin{promptbox}[prompt:expert_evaluator]{Expert Evaluator Task}

You are a rigorous and critical expert evaluator tasked with curating high-quality instruction-following datasets for Large Language Models. Your primary goal is to assess candidate tasks and identify those suitable for a dataset focused on interesting, complex, and multi-step composite instructions.

\vspace{0.5em}
\noindent You will be presented with a candidate task, which includes an instruction (prompt) and several example executions (input/target pairs).

\vspace{0.5em}
\noindent \textbf{Your Objective}:

\noindent Evaluate the candidate task against three critical criteria (detailed below) and decide whether to include it (1) or exclude it (0) from the dataset.

\vspace{0.5em}
\noindent \textbf{Input Format}:

\noindent The task will be provided within the following delimiters:

\vspace{0.5em}
\noindent \textbf{\#\#\# Begin Task Description \#\#\#}

\noindent [Instruction definition]

\noindent \textbf{\#\#\# Example 1 \#\#\#}

\noindent [Input] \dots

\noindent [Target] \dots

\noindent \dots

\noindent \textbf{\#\#\# Example N \#\#\#}

\noindent [Input] \dots

\noindent [Target] \dots

\noindent \textbf{\#\# End Task Description \#\#}

\vspace{1em}
\noindent \textbf{Output Format}:

\noindent You must adhere strictly to the following output format. Do not add any extra text before or after this structure. Failure to comply will cause downstream parsing errors.

\vspace{0.5em}
\noindent [[Rationale]]: ((Your reasoning))

\noindent [[Score]]: ((1 or 0))

\begin{itemize}
    \item \textbf{[[Rationale]]:} Provide a justification for your decision. Be concise but make sure to include relevant details even in the case where the task passes all the relevant details, explain succinctly why. Ensure it is specific enough to justify the result, particularly when detailing errors found in examples. You must explicitly mention which criteria (e.g., "Criterion 1") were decisive in your evaluation.
    \item \textbf{[[Score]]:} Return 0 if the task fails any of the criteria. Return 1 if the task passes all criteria.
\end{itemize}

\vspace{1em}
\noindent \textbf{Evaluation Criteria}

\noindent A superior instruction-following task must satisfy ALL THREE of the following criteria. Evaluate them sequentially.

\vspace{0.5em}
\noindent \textbf{Criterion 1: Absolute Correctness (Critical Veto)}

\noindent The example executions (Targets) must be demonstrably correct and perfectly follow the instruction definition. This is the most important criterion.

\vspace{0.5em}
\noindent Action: Meticulously verify the execution of every example provided.

\noindent Fails if:

\begin{itemize}
    \item ANY example is factually incorrect.
    \item ANY example fails to adhere to the constraints or steps specified in the instruction.
\end{itemize}

\vspace{0.5em}
\noindent \textbf{Example of a Failure (Criterion 1):}

\vspace{0.5em}
\noindent \textbf{\#\#\# Begin Task Description \#\#\#}

\noindent You are given a sentence containing a blank (\_) and two strings, A and B. Your task is to first identify the longest common substring between strings A and B. Then, convert this substring to lowercase and sort its characters alphabetically. Replace the original substring in both A and B with this modified version. Finally, determine which of the modified strings, A or B, logically fills the blank in the sentence based on its content. Your answer should be the modified string (either A or B) that best fits the sentence.

\noindent \textbf{\#\#\# Example 0 \#\#\#}

\noindent [input] Sentence: The \_ was used to build the houses instead of the other material as it was plentiful.

\noindent String A: boardsXYZplentiful

\noindent String B: bricksXYZplentiful

\noindent [target] boardsxyzplentiful

\noindent \textbf{\#\#\# Example 1 \#\#\#}

\noindent [input] Sentence: The insecticide could not kill the insect flying in the house because the \_ is weak.

\noindent String A: insecticideWEAKsubstance

\noindent String B: insectWEAKsubstance

\noindent [target] insecticideaekswubstance

\noindent \textbf{\#\#\# Example 2 \#\#\#}

\noindent [input] Sentence: Tom decided to repaint his bedroom yellow instead of the gray it was now. The \_ is bright.

\noindent String A: yellowBRIGHTcolor

\noindent String B: grayBRIGHTcolor

\noindent [target] yellowbceghilorrt

\noindent \textbf{\#\# End Task Description \#\#}

\vspace{0.5em}
\noindent [[Rationale]]: This task fails Criterion 1 (Absolute Correctness). In Example 0, the instruction requires the longest common substring ("XYZplentiful") to be lowercased and sorted alphabetically. The target "boardsxyzplentiful" shows the substring lowercased but NOT sorted (the sorted version would be "efillnptuxyz"). Example 1 has the same sorting issue. Example 2 is also incorrect: the substring "BRIGHTcolor" sorted alphabetically is "bcghiloorrt", but the target shows "bceghilorrt" (an extra 'e').

\noindent [[Score]]: 0

\vspace{1em}
\noindent \textbf{Criterion 2: Executability and Unambiguity}

\noindent The instruction must be clearly defined, executable, and unambiguous. A reasonably strong language model should be able to execute the task consistently, leading to a single, objective answer.

\vspace{0.5em}
\noindent Action: Analyze the instruction for clarity, subjectivity, and hidden assumptions.

\noindent Fails if:

\begin{itemize}
    \item The task cannot be executed given the information in the prompt, possibly because of too much ambiguity.
    \item The instructions are ambiguous, contradictory, or poorly defined.
    \item The task requires unspecified real-time access or external knowledge that is not provided in the prompt.
\end{itemize}

\vspace{0.5em}
\noindent \textbf{Example of a Failure (Criterion 2):}

\vspace{0.5em}
\noindent \textbf{\#\#\# Begin Task Description \#\#\#}

\noindent Given a command in a limited form of natural language, provide the correct sequence of actions that executes the command to navigate an agent in its environment. Additionally, classify the overall emotion conveyed by the sequence of actions into one of the six emotions: 'joy', 'love', 'anger', 'fear', 'sadness', or 'surprise'.

\vspace{0.5em}
\noindent [\dots instruction details on movement commands omitted for brevity \dots]

\vspace{0.5em}
\noindent The emotion classification should reflect the overall feeling evoked by the agent's movement pattern. For example, rapid, erratic movements might suggest 'fear' or 'anger'\dots

\noindent \textbf{\#\#\# Example 0 \#\#\#}

\noindent [input] run opposite left twice and jump opposite right

\noindent [target] Actions: I\_TURN\_LEFT I\_TURN\_LEFT I\_RUN I\_TURN\_LEFT I\_TURN\_LEFT I\_RUN I\_TURN\_RIGHT I\_TURN\_RIGHT I\_JUMP, Emotion: anger

\noindent \textbf{\#\# End Task Description \#\#}

\vspace{0.5em}
\noindent [[Rationale]]: This task fails Criterion 2 (Executability and Unambiguity). The requirement to classify emotion based on the movement pattern ("The emotion classification should reflect the overall feeling evoked...") is highly subjective. There is no objective standard for determining if the emotion is 'anger' versus 'fear', leading to multiple valid interpretations.

\noindent [[Score]]: 0

\vspace{1em}
\noindent \textbf{Criterion 3: Complexity, Diversity, and Interest}

\noindent The task should be non-trivial, aligning with the goal of collecting complex, multi-step composite instructions. This means the task should require multiple steps or the composition of different skills (e.g., combining math and text analysis, or logic and formatting). Furthermore, the provided examples must be diverse.

\vspace{0.5em}
\noindent Action: Assess the cognitive load required (is it multi-step? does it combine skills?) and the variation across the examples.

\noindent Fails if:

\begin{itemize}
    \item The task is too simple (e.g., basic sentiment classification, simple data extraction).
    \item The task is repetitive or easily gameable (e.g., multiple-choice questions where the answer is always A).
    \item The examples lack diversity in input structure, content, logic, or output format.
\end{itemize}

\vspace{0.5em}
\noindent \textbf{Example of a Pass (All Criteria):}

\vspace{0.5em}
\noindent \textbf{\#\#\# Begin Task Description \#\#\#}

\noindent You are given a snippet of a privacy policy and a multiple-choice question about basic statistics related to data privacy. Your task is to first identify the type of personal information mentioned in the privacy policy snippet. Then, answer the multiple-choice statistics question related to data privacy practices. Provide the identified information type and the letter corresponding to the correct multiple-choice answer.

\noindent \textbf{\#\#\# Example 0 \#\#\#}

\noindent [input] Privacy Policy Snippet: "Our website uses cookies and similar tracking technologies..."

\noindent Statistics Question: A company analyzes website traffic data. They find that the average time spent on their site is 180 seconds with a standard deviation of 30 seconds. Assuming the time spent follows a normal distribution, what percentage of users spend between 150 and 210 seconds on the site?

\noindent (A) 34\% (B) 68\% (C) 95\% (D) 99.7\%

\noindent [target] Information Type: Cookies and tracking elements

\noindent Answer: B

\noindent \textbf{\#\#\# Example 1 \#\#\#}

\noindent [input] Privacy Policy Snippet: "To create an account, we require your email address and a chosen username..."

\noindent Statistics Question: A survey asks users if they are comfortable providing their email address for account creation. 75 out of 100 surveyed users said yes. What is the sample proportion of users comfortable providing their email address?

\noindent (A) 0.25 (B) 0.50 (C) 0.75 (D) 1.00

\noindent [target] Information Type: Contact

\noindent Answer: C

\noindent \textbf{\#\# End Task Description \#\#}

\vspace{0.5em}
\noindent [[Rationale]]: The task passes all criteria. Criterion 1 (Correctness): The solutions to the statistics problems are correct. Criterion 2 (Executability): The instructions are clear and the answers are objective. Criterion 3 (Complexity/Diversity): The task is complex as it requires composing text analysis (privacy policy) with mathematical reasoning (statistics). The examples are diverse (different statistical concepts and variable answer positions).

\noindent [[Score]]: 1

\vspace{1em}
\noindent \textbf{Evaluation Process Checklist}

\noindent When evaluating the task below, strictly follow this sequential process. If a task fails a step, stop evaluating and reject it immediately.

\begin{enumerate}
    \item Verify Correctness (C1): Check every example meticulously. If any fail, reject (0) and explain the specific error.
    \item Assess Executability (C2): If C1 passes, check for ambiguity or subjectivity. If found, reject (0).
    \item Evaluate Complexity and Diversity (C3): If C1 and C2 pass, assess if the task is sufficiently complex and the examples diverse. If not, reject (0).
\end{enumerate}

\noindent Accept (1): Only if the task passes C1, C2, and C3.

\vspace{1em}
\noindent BELOW IS YOUR TASK TO EVALUATE. Remember to strictly follow the output format.

\end{promptbox}

\subsection{Final Dataset Statistics}
\begin{figure}[t!]
\begin{center}
 \includegraphics[scale=0.25]{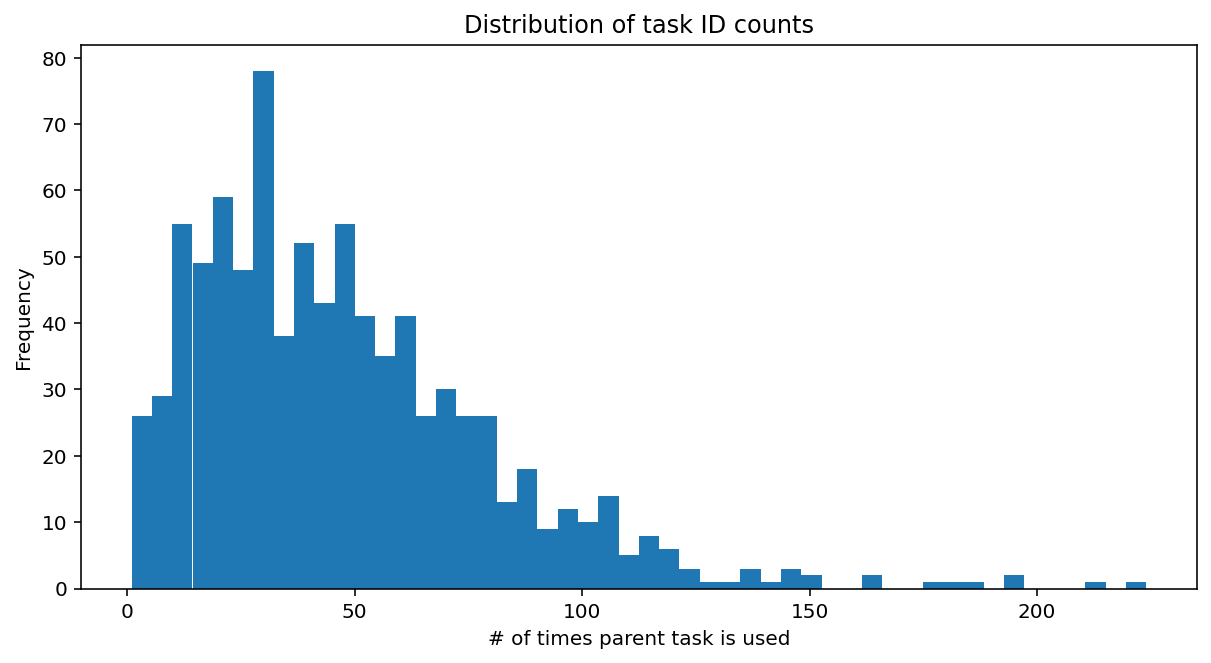}
\caption{Distribution of occurrence statistics of base SNI task appearance in final composite-SNI task list. Some SNI tasks feature in over 200 composite-SNI tasks in the final dataset but most appear in an average of 48 downstream tasks.}
\label{fig:base_sni_occurance}  
\end{center}
\end{figure}

\begin{figure}[t!]
\begin{center}
 \includegraphics[scale=0.25]{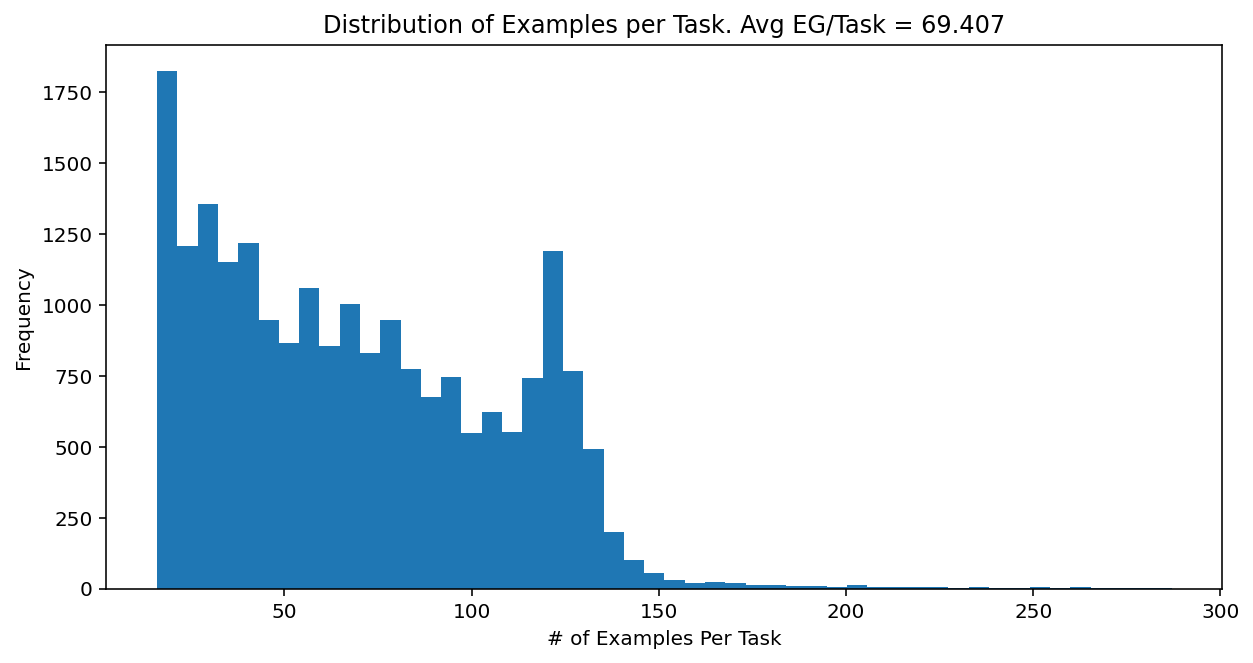}
\caption{A few of the generated tasks have a large number of example instances (> 150). The average number of instances generated per task is $\approx$ 69}
\label{fig:eg_per_task_comp_sni}  
\end{center}
\end{figure}

Figures \ref{fig:base_sni_occurance}, \ref{fig:eg_per_task_comp_sni} show relevant statistics relating to the final Composite SNI dataset.

\subsection{Example Composite SNI Task}

\begin{promptbox}[prompt:cpp_coreference]{[Composite-SNI: $62\_193$] C++ Pronoun Coreference Task}

\textbf{\#\#\# Begin Task Description \#\#\#}

\noindent Given a C++ program and a sentence describing a pronoun coreference relation within it, determine if the reasoning provided for the coreference is correct. The C++ program may contain 'for' loops. Your task is to first count the number of 'for' loops in the provided C++ code snippet. Then, analyze the sentence and the reasoning provided for the pronoun coreference. Finally, output the count of 'for' loops and whether the reasoning is 'Correct' or 'Wrong'.

\vspace{0.5em}
\noindent \textbf{\#\#\# Example Begin \#\#\#}

\noindent \textbackslash{}t[input] C++ Code:

\noindent \texttt{int main() \{}

\noindent \hspace*{2em}\texttt{int a = 1, b = 2, c = 3;}

\noindent \hspace*{2em}\texttt{for (int i = 0; i \textless{} 1; i++) \{}

\noindent \hspace*{4em}\texttt{// He declared multiple variables on one line.}

\noindent \hspace*{2em}\texttt{\}}

\noindent \hspace*{2em}\texttt{return 0;}

\noindent \texttt{\}}

\vspace{0.5em}
\noindent Sentence: He declared multiple variables on one line.

\noindent Reason: The 'He' refers to the programmer because declaration style is a programmer's choice.

\noindent Question: How many 'for' loops are in the code, and is the reasoning correct or wrong?

\noindent [target] 1, Correct

\vspace{0.5em}
\noindent \textbf{\#\#\# Example Begin \#\#\#}

\noindent \textbackslash{}t[input] C++ Code:

\noindent \texttt{int main() \{}

\noindent \hspace*{2em}\texttt{for (int i = 0; i \textless{} 10; i++) \{}

\noindent \hspace*{4em}\texttt{if (i == 5) \{}

\noindent \hspace*{6em}\texttt{break; // He exited the loop when i reached 5.}

\noindent \hspace*{4em}\texttt{\}}

\noindent \hspace*{2em}\texttt{\}}

\noindent \hspace*{2em}\texttt{return 0;}

\noindent \texttt{\}}

\vspace{0.5em}
\noindent Sentence: He exited the loop when i reached 5.

\noindent Reason: The 'He' refers to the 'break' statement because 'break' causes the exit.

\noindent Question: How many 'for' loops are in the code, and is the reasoning correct or wrong?

\noindent [target] 1, Wrong

\vspace{0.5em}
\noindent \textbf{\#\#\# Example Begin \#\#\#}

\noindent \textbackslash{}t[input] C++ Code:

\noindent \texttt{int main() \{}

\noindent \hspace*{2em}\texttt{int x = 10;}

\noindent \hspace*{2em}\texttt{int y = x / 2;}

\noindent \hspace*{2em}\texttt{// It holds half the value of x.}

\noindent \hspace*{2em}\texttt{return y;}

\noindent \texttt{\}}

\vspace{0.5em}
\noindent Sentence: It holds half the value of x.

\noindent Reason: The 'It' refers to the variable 'y' because 'y' is assigned the result of the division.

\noindent Question: How many 'for' loops are in the code, and is the reasoning correct or wrong?

\noindent [target] 0, Correct

\end{promptbox}

The above task was constructed from the following source tasks derived from the SNI dataset:

\begin{promptbox}[prompt:cpp_for_loop]{[SNI Task: 62] C++ For Loop Counting Task}

\textbf{\#\#\# Begin Task Description \#\#\#}

\noindent This task is to find the number of 'For' loops present in the given cpp program.

\vspace{0.5em}
\noindent \textbf{\#\#\# Example Begin \#\#\#}

\noindent \textbackslash{}t[input]\texttt{int turn(int a,int b)}

\noindent \texttt{\{}

\noindent \hspace*{2em}\texttt{int i,k;}

\noindent \hspace*{2em}\texttt{if(a==1) k=1;}

\noindent \hspace*{2em}\texttt{else if(a\textless{}b) k=0;}

\noindent \hspace*{2em}\texttt{else}

\noindent \hspace*{2em}\texttt{\{}

\noindent \hspace*{4em}\texttt{k=0;}

\noindent \hspace*{4em}\texttt{for(i=b;i\textless{}=a;i++)}

\noindent \hspace*{6em}\texttt{if(a\%i==0)}

\noindent \hspace*{8em}\texttt{k=k+turn(a/i,i);}

\noindent \hspace*{2em}\texttt{\}}

\noindent \hspace*{2em}\texttt{return k;}

\noindent \texttt{\}}

\noindent \texttt{int main()}

\noindent \texttt{\{}

\noindent \hspace*{2em}\texttt{int n,i,a[1000],t;}

\noindent \hspace*{2em}\texttt{scanf("\%d",\&n);}

\noindent \hspace*{2em}\texttt{for(i=0;i\textless{}n;i++)}

\noindent \hspace*{2em}\texttt{\{}

\noindent \hspace*{4em}\texttt{scanf("\%d",\&a[i]);}

\noindent \hspace*{4em}\texttt{t=turn(a[i],2);}

\noindent \hspace*{4em}\texttt{printf("\%d\textbackslash{}n",t);}

\noindent \hspace*{2em}\texttt{\}}

\noindent \hspace*{2em}\texttt{return 0;}

\noindent \texttt{\}}

\vspace{0.5em}
\noindent [target] 2

\vspace{0.5em}
\noindent \textbf{\#\#\# Example Begin \#\#\#}

\noindent \textbackslash{}t[input]\texttt{int devide(int n,int k)}

\noindent \texttt{\{}

\noindent \hspace*{2em}\texttt{int s,i;}

\noindent \hspace*{2em}\texttt{s=1;}

\noindent \hspace*{2em}\texttt{for(i=k;i\textless{}=sqrt(n*1.0);i++)}

\noindent \hspace*{2em}\texttt{\{}

\noindent \hspace*{4em}\texttt{if(n\%i==0)}

\noindent \hspace*{4em}\texttt{\{}

\noindent \hspace*{6em}\texttt{s=s+devide(n/i,i);}

\noindent \hspace*{4em}\texttt{\}}

\noindent \hspace*{2em}\texttt{\}}

\noindent \hspace*{2em}\texttt{return s;}

\noindent \texttt{\}}

\noindent \texttt{int main()}

\noindent \texttt{\{}

\noindent \hspace*{2em}\texttt{int n,a[200],i;}

\noindent \hspace*{2em}\texttt{scanf("\%d",\&n);}

\noindent \hspace*{2em}\texttt{for(i=0;i\textless{}n;i++)}

\noindent \hspace*{2em}\texttt{\{}

\noindent \hspace*{4em}\texttt{scanf("\%d",\&a[i]);}

\noindent \hspace*{4em}\texttt{if(i==0)printf("\%d",devide(a[i],2));}

\noindent \hspace*{4em}\texttt{else printf("\textbackslash{}n\%d",devide(a[i],2));}

\noindent \hspace*{2em}\texttt{\}}

\noindent \hspace*{2em}\texttt{return 0;}

\noindent \texttt{\}}

\vspace{0.5em}
\noindent [target] 2

\end{promptbox}

\begin{promptbox}[prompt:coreference_reasoning]{[SNI Task 193] Pronoun Coreference Reasoning Task}

\textbf{\#\#\# Begin Task Description \#\#\#}

\noindent In this task you need to indicate the plausibility of reasoning for the pronoun coreference relations. Each of the provided inputs contains a sentence with a target pronoun and a sentence that justifies which noun phrase the pronoun refers to. Correct reasons do not need to use all the knowledge from the sentence. The resolution of the pronoun coreference relations typically involve one or multiple following knowledge types about commonsense: First: 'Property', the knowledge about property of objects (e.g., ice is cold). Second: 'Object', the knowledge about objects (e.g., cats have ears). Third: 'Eventuality', the knowledge about eventuality (e.g., 'wake up' happens before 'open eyes'). Fourth: 'Spatial', the knowledge about spatial position (e.g., object at the back can be blocked). Fifth: 'Quantity', the knowledge about numbers (e.g., 2 is smaller than 10). Sixth: all other knowledge if above ones are not suitable. You should answer 'Correct' if the reasoning made sense, otherwise, you should answer 'Wrong'.

\vspace{0.5em}
\noindent \textbf{\#\#\# Example Begin \#\#\#}

\noindent \textbackslash{}t[input]Sentence: Jackson was greatly influenced by Arnold, though he lived two centuries earlier.

\noindent Reason: The `he' refers to arnold because ``Was'' is past tense. He was influenced by Arnold. So you assume Arnold is dead. 

\noindent Question: Is the above reasoning correct or wrong? 

\noindent [target]Correct

\vspace{0.5em}
\noindent \textbf{\#\#\# Example Begin \#\#\#}

\noindent \textbackslash{}t[input]Sentence: I put the heavy book on the table and it broke.

\noindent Reason: The `it' refers to the table because A butterfly wing would likely not be heavy enough to break a table. 

\noindent Question: Is the above reasoning correct or wrong? 

\noindent [target]Wrong

\vspace{0.5em}
\noindent \textbf{\#\#\# Example Begin \#\#\#}

\noindent \textbackslash{}t[input]Sentence: The sculpture rolled off the shelf because it wasn't level.

\noindent Reason: The `it' refers to the shelf because The object not anchored is the one likely to roll. 

\noindent Question: Is the above reasoning correct or wrong? 

\noindent [target]Wrong

\end{promptbox}

\section{Retrieval Details}\label{appendix:retrieval-details}
Given a library $\mathcal{T}_{src}[T] = {T_1, \ldots, T_N}$ of $N$ source tasks, the module retriever identifies which subset of tasks is most relevant to a novel query task $T$. 

\paragraph{Task Embedding.} Each source task $T_i$ in the library is associated with a training dataset $\mathcal{D}_i$. To obtain a fixed-dimensional representation of a task, we sample $K$ training examples from $\mathcal{D}_i$ and embed each example independently using a pretrained text embedding model. Concretely, for each example $x_j^{(i)} \in \mathcal{D}_i$, we compute $\mathbf{e}_j^{(i)} = \texttt{Embed}(x_j^{(i)}) \in \mathbb{R}^d$, yielding an embedding table $E_i = [\mathbf{e}_1^{(i)}, \ldots, \mathbf{e}_K^{(i)}] \in \mathbb{R}^{K \times d}$ for each task. In our experiments, we use $K{=}16$ training examples per task and the Gemini Embedding model (gemini-embedding-001) \citep{gemini-embedding} with the classification task type.

\paragraph{Retriever Architecture.} We frame task retrieval as an $N$-way classification problem over the source tasks. The retriever is a lightweight multilayer perceptron (MLP) $f_\theta: \mathbb{R}^d \to \mathbb{R}^N$ with hidden layers of sizes $(512, 256)$ and ReLU activations. Each class corresponds to one task $m_i$ in the source-task library. The MLP is trained on the union of all per-task embedding tables:

$$\mathcal{T} = \bigcup_{i=1}^{N} \{(\mathbf{e}_j^{(i)}, i) \mid j=1,\ldots,K\}$$

\noindent using the standard softmax cross-entropy loss where the input is the embedding $e_{j} \in \mathbb{R}^d$ and the label to predict is $i$, the unique task identifier. 

The training data is split 80/20 into train and validation sets. We optimize with Adam (learning rate $10^{-3}$, batch size 128) for up to 1000 epochs with early stopping. 

\paragraph{Inference: Ensemble Scoring.}
At inference time, given a query task with $K$ embedded examples $E_q = [\mathbf{e}_1^{(q)}, \ldots, \mathbf{e}_K^{(q)}]$, we compute an ensemble score for each source task in the library by aggregating the per-example log-probabilities:

$$s_i = \sum_{j=1}^{K} \log \text{softmax}(f_\theta(\mathbf{e}_j^{(q)}))_i$$

\noindent The raw scores are then normalized via z-scoring and a $\tanh$ squashing function:

$$\hat{s}_i = \tanh\left(\frac{s_i - \mu_s}{\sigma_s}\right)$$

\noindent where $\mu_s$ and $\sigma_s$ are the mean and standard deviation of the raw scores across all $N$ tasks. Source tasks are ranked by $\hat{s}_i$ in descending order, and the tasks with the top-$k$ scores are selected for downstream use.

\section{LLM Pair Selection Details}\label{appendix:llm-selection}
Given $\tau$ and $N / 2$ source-task pairs as given by the retriever, we use Gemini 2.5 to compare these pairs with $\tau$ and output the most relevant task pair. Specifically, each source task $T_i$ (and also $\tau)$ is associated with a \textit{task description} comprising a natural-language task definition and $E$ input-output demonstration examples drawn from the task's training data. These are formatted using delimiter tags:

\begin{lstlisting}
<instruction> [task definition] </instruction>
<example> [input_1] [target_1] </example>
<example> [input_2] [target_2] </example>
...
<example> [input_E] [target_E] </example>
\end{lstlisting}

In our experiments, we use $E = 2$ examples. Next, we ask an LLM, in our case Gemini 2.5 Pro, to process the task descriptions for the candidate pairs and the target task and to output the most relevant candidate pair along with a rationale:

\begin{promptbox}[prompt:pair_ranking]{Pair Ranking for Target Task}\label{prompt:pair-ranking}
\ttfamily
I have a target task described below:\\
Target Task: \{task\_input\}

I have a list of candidate pairs of other tasks. For each pair, consider if knowing how to solve these two tasks would help solve the target task.\\
Candidate Pairs:\\
\{candidate\_pairs\}

Which pair is the MOST useful for the target task? Output the index of the pair (0-indexed) and a brief reason.\\
Format: Pair Index: <index>\\
Reason: <reason>

\end{promptbox}

\section{Source Task Sensitivity Results}
\label{appendix:source_task_sensitivity}
\begin{table}[htbp]
\centering
\caption{Task Performance Comparison (Rounded).}
\label{tab:performance_data}
\resizebox{\textwidth}{!}{
\begin{tabular}{|l|c|c|c|c|}
\hline
\textbf{Task}&\textbf{True Source Tasks}&\textbf{Retrieval + LLM Selection}&\textbf{Random + LLM Selection}&\textbf{Direct Training} \\ \hline
\textbf{100\_870}&\textcolor{blue}{0.5649}&0.6332&0.6219&\textcolor{red}{0.7053} \\ \hline
\textbf{101\_269}&\textcolor{blue}{1.0197}&1.0841&1.0682&\textcolor{red}{1.3763} \\ \hline
\textbf{1\_450} & \textcolor{blue}{0.1834} & 0.2428 & 0.2320 & \textcolor{red}{0.3793} \\ \hline
\textbf{31\_91} & \textcolor{blue}{0.4099} & 0.4964 & 0.5707 & \textcolor{red}{0.7620} \\ \hline
\textbf{330\_736} & 0.0999 & \textcolor{blue}{0.0000} & \textcolor{blue}{0.0000} & \textcolor{red}{0.0106} \\ \hline
\textbf{41\_174} & \textcolor{blue}{0.0999} & 0.3438 & 0.4096 & \textcolor{red}{0.6084} \\ \hline
\textbf{43\_122} & 1.1463 & \textcolor{blue}{0.9278} & 1.0524 & \textcolor{red}{1.6496} \\ \hline
\textbf{583\_850} & \textcolor{blue}{0.3461} & 0.4149 & 0.4055 & \textcolor{red}{0.5242} \\ \hline
\textbf{625\_640} & \textcolor{blue}{0.0000} & 0.1328 & \textcolor{red}{0.2653} & 0.1368 \\ \hline
\textbf{667\_794} & \textcolor{blue}{0.0507} & 0.0789 & \textcolor{red}{0.1176} & 0.0920 \\ \hline
\textbf{79\_411} & 0.2708 & \textcolor{blue}{0.0792} & \textcolor{red}{0.1140} & 0.0882 \\ \hline
\textbf{869\_873} & \textcolor{blue}{0.6440} & 0.6455 & 0.6450 & \textcolor{red}{0.7729} \\ \hline
\textbf{91\_595} & 0.9210 & \textcolor{blue}{0.9208} & 0.9221 & \textcolor{red}{1.0107} \\ \hline
\textbf{96\_721} & 0.0030 & 0.0007 & \textcolor{blue}{0.0000} & \textcolor{red}{0.0274} \\ \hline
\textbf{96\_827} & \textcolor{blue}{1.0050} & 1.0216 & 1.0210 & \textcolor{red}{1.1932} \\ \hline
\end{tabular}%
}
\end{table}

\subsection{Ablation: Sensitivity to Source Task Selection}\label{sec:composer-noisy}
Having established the heuristic task selection pipeline in \Cref{subsubsection:task-retrieval}, which we refer to \textbf{Retrieval + LLM Selection}, we now investigate \tempname's sensitivity to the quality of these retrieved source sets by investigating \textbf{Random + LLM Selection} -- a setting where we replace the initial retrieval phase with a random sampling of task pairs. We compare these two heuristic approaches against both an upper bound (using the true source tasks from the base SNI dataset) and a standard direct prefix-tuning baseline.

\begin{figure}[h!]
\centering
\includegraphics[width=\linewidth, height=7cm, keepaspectratio]{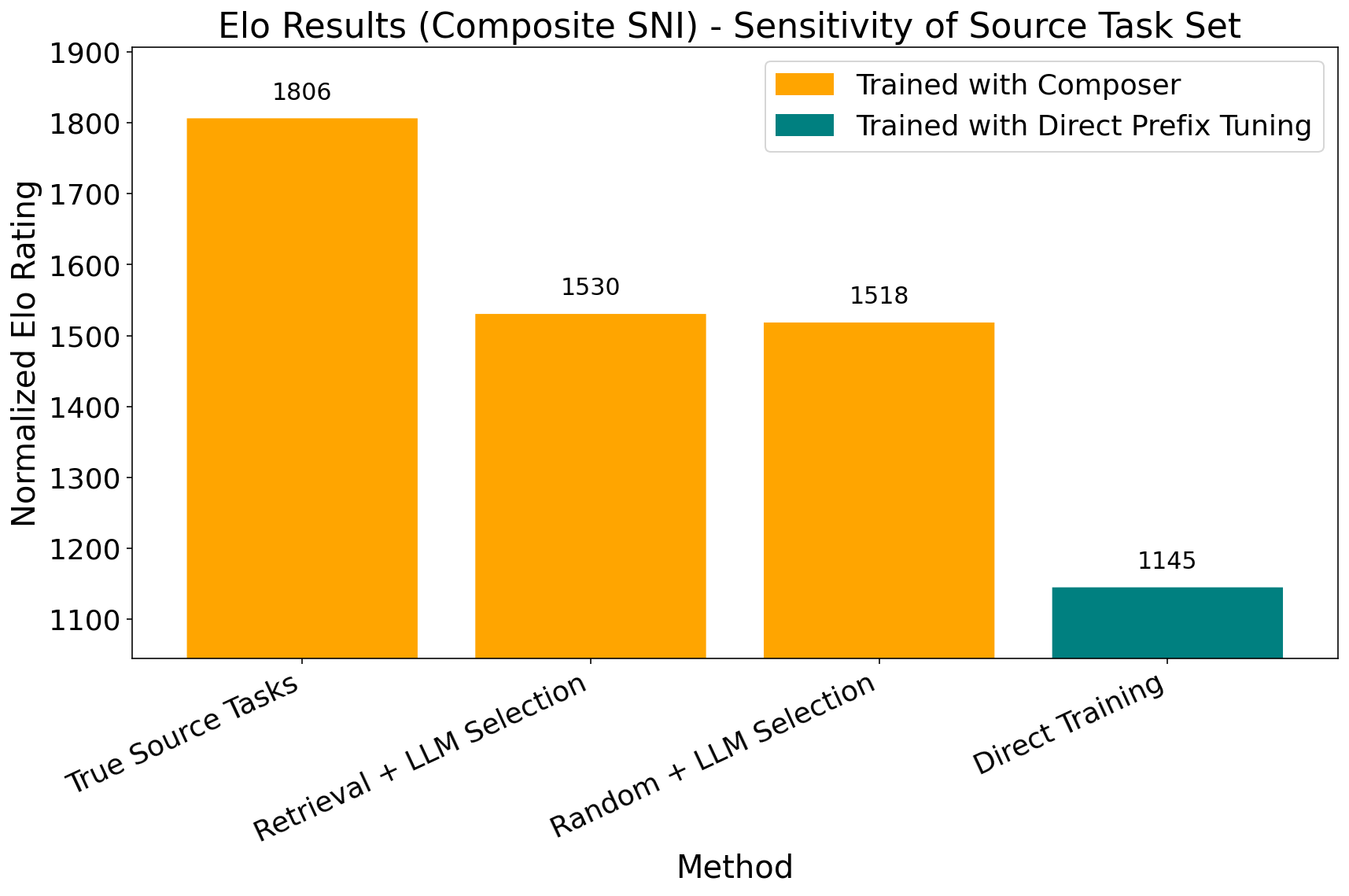}
\caption{Our results demonstrate that even in the absence of ground-truth, synthesizing an intelligently selected set of source tasks can boost performance on target tasks. (Elo calculated over performance on 15 tasks)}
\label{fig:sensitivity_to_source}
\end{figure}

\looseness-1 Our findings (\Cref{fig:sensitivity_to_source}) demonstrate that \tempname maintains a substantial performance advantage over the direct training baseline, even when operating without ground-truth knowledge. While utilizing the true source tasks predictably yields the strongest overall performance, synthesizing tasks selected via heuristic pipelines still confers a significant empirical advantage over learning from scratch.

Though the Elo scores show that \textit{Retrieval + LLM Selection} is only \textit{slightly} better than its random counterpart, a deeper dive into the per-task performance (see above \Cref{tab:performance_data}) reveals that whilst \textit{Random + LLM Selection} can result in performance that is significantly worse than Direct Training (3/15 tasks), the performance of \textit{Retrieval + LLM Selection} is superior to the baseline across all 15 tasks. This suggests future work on improving the retrieval mechanism.

\section{Rationale Prompt}
\label{appendix:rationale_prompt}
\begin{promptbox}[prompt:rationale_prompt]{Rationale Prompt}
\ttfamily
Target Task: \{task\_input\}

Selection:
\{selected\_pair\}

Please write a detailed rationale for why solving the selected pair of tasks (independently or jointly) might result in useful skills/knowledge transfer to the target task. If any or none of the pair of tasks has transfer value for the target, then indicate so. Do not mention the task IDs of the selected pair number in the final rationale, just the task details. The rationale should start with 'Rationale: '.
\end{promptbox}

\section{\tempname Training Details}
\label{appendix:training_composer}
For training \tempname, we use the official pre-trained Gemma-4B \citep{gemmateam2025gemma3technicalreport} as the augmented co-processor model. We use the Adam optimizer \citep{kingma2014adam} with learning rates $\{1e^{-3}, 1e^{-4}\}$ and batch sizes $\{32, 64, 128\}$. For experiments with Composite-SNI, we train for 10000 steps whilst we use 2000 steps when training for the base SNI tasks. To reduce overfitting, we introduce data augmentations during training where parts of the model input are occasionally dropped out. We drop-out KV-caches by setting the whole cache to zero. We explore KV dropout probabilities in the set $\{0.1, 0.293, 0.5\}$. During training, we allow the composer to randomly sample the output length of the generated cache from the set $\{16, 32, 64, 128\}$, but during evaluation, we always output caches of length 32.

\subsubsection{\tempname Training data example}
\label{appendix:training_composer_td_eg}
Below is the full sequence of a single training example from the SNI Meta-Training set.
\begin{promptbox}[prompt:preamble-text]{Preamble Text}
\label{prompt:preamble_text}
\ttfamily
Your role is Cognitive Grafting. Take the established root systems (K-V prefixes) from the varied source tasks and graft them together to support the growth of the new, novel downstream task. Next is information about the source 
tasks
\end{promptbox}

\begin{promptbox}[prompt:source-text-1]{Source Text 1}
\label{prompt:preamble_text-1}
\ttfamily
</instruction> You are given a question on professional accounting. You are also given 4 answer options (associated with ``A'', ``B'', ``C'', ``D''), out of which only one is correct. You need to answer the question by selecting
the correct option. You should only answer with the choice letter, not the whole answer. </instruction>"
\end{promptbox}

\begin{promptbox}[prompt:source-text-2]{Source Text 2}
\label{prompt:preamble_text-2}
\ttfamily
</instruction> You are provided with a user review of a restaurant. Your task is to classify the given review into two categories: 1) positive, and 2) negative based on its content. </instruction>
\end{promptbox}

\begin{promptbox}[prompt:combo-text]{Combination Text}
\label{prompt:combo_text}
\ttfamily
\noindent The preceding source tasks were combined using the following rationale:

\vspace{0.5em}
\noindent <rationale>
The target task involves analyzing a product review to determine its sentiment (positive or negative) and comparing that sentiment to a given numerical rating. The core required skill is text sentiment analysis to determine if the review implies a positive or negative experience.

The selected pair includes a task (Task 298) that requires classifying restaurant reviews into 'positive' or 'negative' categories. This directly aligns with the primary skill needed for the target task: interpreting subjective text to extract a binary sentiment. The model trained on this task learns to identify keywords, phrases, and overall tone that indicate satisfaction or dissatisfaction. This skill is crucial for evaluating whether the sentiment expressed in the review matches the numerical rating provided in the target task input.

The other task in the pair (Task 828) is an unrelated accounting multiple-choice question task, which does not offer much relevant skill transfer. However, Task 298 provides a strong, direct match for the core sentiment analysis component, making this pair a suitable choice for skill transfer to the target task.
</rationale>

\vspace{0.5em}
\noindent to create a new task, which you will be asked to solve. 

\vspace{1.5em}
\noindent Here are a few examples of the task you will be solving:

\vspace{1em}
\noindent \textbackslash{}n<example>
In this task, you're given a review from Amazon and rating for the product on a scale of 1-5 based on the review. The rating means 1: extremely poor, 2: poor, 3: neutral, 4: good, 5: extremely good. Your task is to generate whether the rating matches the review or not. Answer with ``True'' if the sentence belongs to that section, otherwise answer with ``False''.
\vspace{0.5em}
\noindent \textbackslash{}t [input]Review: As soon as I opened it, I didn’t like it because the product looks old and used and rusted. I decided to give it a try but it took like half an hour to heat up. In addition to it, I don’t like the curling iron. It doesn’t curl at all. Straightener is good though. I am returning the item.
\noindent \textbackslash{}n Rating: 1
\noindent \textbackslash{}n [target]True
\noindent \textbackslash{}n </example>

\vspace{1em}
\noindent \textbackslash{}n<example>
In this task, you're given a review from Amazon and rating for the product on a scale of 1-5 based on the review. The rating means 1: extremely poor, 2: poor, 3: neutral, 4: good, 5: extremely good. Your task is to generate whether the rating matches the review or not. Answer with ``True'' if the sentence belongs to that section, otherwise answer with ``False''.
\vspace{0.5em}
\noindent \textbackslash{}t [input]Review: I bought this awhile back. This mattress is quite firm so if you aren't looking for that, I would probably avoid this mattress. It expanded just as expected and was definitely a great mattress considering how cheap it was.
\noindent \textbackslash{}n Rating: 2
\noindent \textbackslash{}n [target]False
\noindent \textbackslash{}n </example>

\vspace{1em}
\noindent \textbackslash{}n<example>
In this task, you're given a review from Amazon and rating for the product on a scale of 1-5 based on the review. The rating means 1: extremely poor, 2: poor, 3: neutral, 4: good, 5: extremely good. Your task is to generate whether the rating matches the review or not. Answer with ``True'' if the sentence belongs to that section, otherwise answer with ``False''.
\vspace{0.5em}
\noindent \textbackslash{}t [input]Review: I purchased 2 bookmarks and received them yesterday. Both screens don't work. I was about to return one, but now the other one said it is ineligible for a return/replacement. Not impressed.
\noindent \textbackslash{}n Rating: 1
\noindent \textbackslash{}n [target]True
\noindent \textbackslash{}n </example>

\vspace{1.5em}
\noindent Now please generate the K-V cache that can be used to solve this given task when given new unseen test examples.
\end{promptbox}

\section{Raw NLL Scores}\label{appendix:raw_nll_scores}

\begin{table*}[htbp]
\centering
\caption{Raw Negative Log-Likelihood (NLL) on Composite-SNI: Zero-Shot, ICL, and Weight-Space Initializations. (Lower is better).}
\label{tab:raw_nll_zeroshot}
\small
\resizebox{\textwidth}{!}{%
\begin{tabular}{l|ccccccc|cc}
\hline
\textbf{Task ID} & \textbf{Parent Avg} & \textbf{Concat} & \textbf{LERP} & \textbf{SVD-Embed} & \textbf{SVD-Head} & \textbf{SVD-Seq} & \textbf{ICL-Init ZS} & \textbf{ICL} & \textbf{SkillSmith (ZS)} \\ \hline
\textbf{1\_450} & 2.5800 & 2.6200 & 2.6284 & 2.5910 & 2.6128 & 2.6070 & 2.6480 & 0.4416 & 0.8708 \\
\textbf{31\_91} & 2.1100 & 2.0188 & 1.9513 & 2.0885 & 2.1634 & 2.1673 & 2.1990 & 0.8295 & 0.6362 \\
\textbf{41\_174} & 6.3605 & 6.4461 & 6.5336 & 9.0098 & 8.2121 & 8.0883 & 5.2100 & 0.8580 & 1.5269 \\
\textbf{43\_122}& 2.1207 & 2.1464 & 2.1074 & 2.1104 & 2.0985 & 2.1047 & 2.1400 & 1.0430 & 1.7436 \\
\textbf{79\_411} & 11.5230 & 11.2963 & 10.6155 & 13.8411 & 11.6429 & 11.9085 & 11.6800 & 1.9610 & 0.1968 \\
\textbf{91\_595} & 1.8326 & 1.7886 & 1.7988 & 1.8573 & 1.7957 & 1.7943 & 1.8130 & 0.9901 & 1.0572 \\
\textbf{96\_721}& 2.9530 & 3.0372 & 2.9455 & 3.3154 & 3.3144 & 3.3138 & 3.2960 & 0.1460 & 0.9202 \\
\textbf{96\_827}  & 2.6139 & 2.6399 & 2.6034 & 2.5904 & 2.5990 & 2.5983 & 2.6360 & 1.0010 & 1.6001 \\
\textbf{100\_870} & 3.1702 & 3.1399 & 2.9852 & 3.2037 & 3.1432 & 3.1193 & 2.6080 & 0.7800 & 1.3528 \\
\textbf{101\_269} & 3.0830 & 3.0477 & 3.0547 & 3.1226 & 3.1057 & 3.1053 & 3.1970 & 1.1936 & 1.8002 \\
\textbf{330\_736} & 8.1049 & 7.6368 & 8.4033 & 9.1675 & 9.1498 & 9.0877 & 8.6360 & 1.5540 & 0.4654 \\
\textbf{583\_850} & 4.0360 & 4.0434 & 3.6181 & 3.5950 & 3.5448 & 3.5423 & 3.3190 & 0.4490 & 3.0320 \\
\textbf{667\_794} & 6.4471 & 6.3629 & 6.1754 & 6.3289 & 6.3341 & 6.3613 & 2.7670 & 0.2770 & 0.8779 \\
\textbf{869\_873} & 1.2255 & 1.1658 & 0.9661 & 0.9933 & 0.9443 & 0.9409 & 0.9510 & 0.6847 & 0.6881 \\
\textbf{625\_640} & 11.0340 & 10.9995 & 11.0538 & 11.1835 & 11.0508 & 10.9446 & 10.6950 & 1.6160 & 0.3733 \\ \hline
\end{tabular}%
}
\end{table*}

\begin{table*}[htbp]
\centering
\caption{Raw Negative Log-Likelihood (NLL) on Composite-SNI: Performance Following Fine-Tuning. (Lower is better).}
\label{tab:raw_nll_finetuned}
\small
\resizebox{\textwidth}{!}{%
\begin{tabular}{l|c|ccccccc|cc}
\hline
\textbf{Task ID} & \textbf{Direct Tr.} & \textbf{Parent Avg} & \textbf{Concat} & \textbf{LERP} & \textbf{SVD-Embed} & \textbf{SVD-Head} & \textbf{SVD-Seq} & \textbf{ICL-Init FT} & \textbf{SkillSmith (True Sources)} \\ \hline
\textbf{1\_450} & 0.3793 & 0.2950 & 0.2974 & 0.3356 & 0.2620 & 0.3073 & 0.3243 & 0.2920 & 0.1834 \\
\textbf{31\_91} & 0.7620 & 0.8750 & 0.8139 & 0.7226 & 0.7982 & 0.8116 & 0.8391 & 0.7060 & 0.4099 \\
\textbf{41\_174}  & 0.6084 & 0.5411 & 0.7213 & 0.5339 & 0.4223 & 0.3242 & 0.3538 & 0.1760 & 0.0999 \\
\textbf{43\_122}  & 1.6496 & 1.6659 & 1.8626 & 1.6357 & 1.6257 & 1.6440 & 1.5945 & 1.2620 & 1.1463 \\
\textbf{79\_411}  & 0.0882 & 0.3195 & 0.6856 & 0.5381 & 0.3677 & 0.2070 & 0.3175 & 1.3250 & 0.2708 \\
\textbf{91\_595}  & 1.0107 & 1.0608 & 1.0162 & 1.0195 & 1.0456 & 1.0549 & 1.0707 & 1.0230 & 0.9210 \\
\textbf{96\_721}  & 0.0274 & 0.0309 & 0.0253 & 0.0235 & 0.0218 & 0.0334 & 0.0182 & 0.0180 & 0.0030 \\
\textbf{96\_827}  & 1.1932 & 1.2045 & 1.2621 & 1.2105 & 1.1520 & 1.2037 & 1.0914 & 1.1090 & 1.0050 \\
\textbf{100\_870} & 0.7053 & 0.7109 & 0.7358 & 0.7552 & 0.7235 & 0.6411 & 0.6550 & 0.5860 & 0.5649 \\
\textbf{101\_269} & 1.3763 & 1.3869 & 1.3669 & 1.3572 & 1.3612 & 1.3496 & 1.3547 & 1.2080 & 1.0197 \\
\textbf{330\_736} & 0.0106 & 0.3231 & 0.2034 & 0.4816 & 0.2841 & 0.4134 & 0.3136 & 0.0470 & 0.0999 \\
\textbf{583\_850} & 0.5242 & 0.3973 & 0.4993 & 0.4285 & 0.4285 & 0.4052 & 0.4366 & 0.4870 & 0.3461 \\
\textbf{667\_794} & 0.0920 & 0.1029 & 0.1543 & 0.1376 & 0.1732 & 0.0821 & 0.1928 & 0.0820 & 0.0507 \\
\textbf{869\_873} & 0.7729 & 0.7936 & 0.8243 & 0.7930 & 0.7996 & 0.8296 & 0.8101 & 0.7490 & 0.6440 \\
\textbf{625\_640} & 0.1368 & 0.0887 & 0.0570 & 0.1229 & 0.1493 & 0.2155 & 0.0438 & 0.0450 & 0.0000 \\ \hline
\end{tabular}%
}
\end{table*}

\begin{table*}[htbp!]
\centering
\caption{Raw Negative Log-Likelihood (NLL) on Super-Natural Instructions (SNI): Zero-Shot, ICL, and Initializations. (Lower is better).}
\label{tab:sni_nll_zeroshot}
\small
\resizebox{\textwidth}{!}{%
\footnotesize
\setlength{\tabcolsep}{3pt}
\renewcommand{\arraystretch}{0.95}
\begin{tabular}{l|ccccccc|c|cc}
\hline
\textbf{Task} & \textbf{Parent} & \textbf{Concat} & \textbf{LERP} & \textbf{SVD} & \textbf{SVD} & \textbf{SVD} & \textbf{ICL-Init} & \textbf{ICL} & \textbf{SkillSmith} & \textbf{SkillSmith} \\
\textbf{ID} & \textbf{Avg} & & & \textbf{Embed} & \textbf{Head} & \textbf{Seq} & \textbf{ZS} & & \textbf{(Retrieved)} & \textbf{(CSNI-Pre)} \\ \hline
\textbf{57}  & 7.5625  & 7.6265  & 8.0553  & 8.1522  & 8.2454  & 8.1967  & 8.0772  & 2.0198 & 0.6252 & 0.3985 \\
\textbf{133} & 3.5278  & 3.5995  & 3.5273  & 3.5164  & 3.4602  & 3.4947  & 3.4990  & 1.0906 & 0.3078 & 0.1992 \\
\textbf{171} & 2.0432  & 1.8293  & 1.8325  & 2.5682  & 1.9435  & 1.9337  & 1.9817  & 0.9786 & 1.2097 & 1.0847 \\
\textbf{204} & 11.3479 & 11.3024 & 11.1291 & 11.6804 & 11.6542 & 11.5744 & 7.2891  & 2.2348 & 0.3235 & 0.3519 \\
\textbf{393} & 4.2804  & 4.4789  & 3.4612  & 3.7875  & 3.5375  & 3.5528  & 3.8527  & 1.1926 & 1.9743 & 1.3607 \\
\textbf{549} & 4.6227  & 4.5129  & 4.7049  & 5.7759  & 5.1111  & 5.0962  & 5.1828  & 1.3415 & 1.6529 & 0.3667 \\
\textbf{691} & 8.1169  & 7.8365  & 6.7845  & 7.0051  & 7.5992  & 7.5915  & 7.5849  & 2.1524 & 5.5604 & 3.3689 \\
\textbf{731} & 10.6090 & 11.2829 & 10.2385 & 11.3719 & 11.8686 & 11.9030 & 9.3789  & 1.3625 & 0.5422 & 0.5549 \\
\textbf{780} & 6.8185  & 6.6995  & 5.5661  & 5.8363  & 6.4860  & 6.4141  & 6.4399  & 1.3462 & 3.4754 & 3.5575 \\
\textbf{843} & 10.0737 & 9.7736  & 10.1783 & 10.2499 & 10.2811 & 10.1875 & 10.2819 & 3.1265 & 3.6243 & 0.9736 \\ \hline
\end{tabular}
}
\end{table*}

\begin{table*}[htbp!]
\centering
\caption{Raw Negative Log-Likelihood (NLL) on Super-Natural Instructions (SNI): Performance Following Downstream Fine-Tuning. (Lower is better).}
\label{tab:sni_nll_finetuned}
\small
\resizebox{\textwidth}{!}{%
\footnotesize
\setlength{\tabcolsep}{4.5pt}
\renewcommand{\arraystretch}{0.95}
\begin{tabular}{l|c|cccccc|c|ccc}
\hline
\textbf{Task} & \textbf{Direct} & \textbf{Parent} & \textbf{Concat} & \textbf{LERP} & \textbf{SVD} & \textbf{SVD} & \textbf{SVD} & \textbf{ICL-Init} & \textbf{SkillSmith} & \textbf{SkillSmith} & \textbf{SkillSmith} \\
\textbf{ID} & \textbf{Tr.} & \textbf{Avg} & & & \textbf{Embed} & \textbf{Head} & \textbf{Seq} & \textbf{FT} & \textbf{(Retrieved)} & \textbf{(CSNI-Pre)} & \textbf{(CSNI-Pre-Ret)} \\ \hline
\textbf{57}  & 0.3478 & 0.3060 & 0.3401 & 0.2574 & 0.2994 & 0.2762 & 0.1753 & 0.3324 & 0.3349 & 0.1585 & 0.1869 \\
\textbf{133} & 0.0226 & 0.0078 & 0.0060 & 0.0122 & 0.0044 & 0.0089 & 0.0079 & 0.0027 & 0.0036 & 0.0114 & 0.0038 \\
\textbf{171} & 0.7604 & 0.6776 & 0.6442 & 0.6489 & 0.6899 & 0.6857 & 0.6999 & 0.6296 & 0.6875 & 0.6631 & 0.6366 \\
\textbf{204} & 0.1134 & 0.0350 & 0.0268 & 0.1236 & 0.0430 & 0.0249 & 0.0121 & 0.0449 & 0.0395 & 0.0300 & 0.0435 \\
\textbf{393} & 0.2134 & 0.1525 & 0.1725 & 0.1396 & 0.1526 & 0.1646 & 0.1606 & 0.1401 & 0.1441 & 0.1382 & 0.1696 \\
\textbf{549} & 0.2405 & 0.1750 & 0.1533 & 0.1945 & 0.1280 & 0.1508 & 0.1967 & 0.1079 & 0.1726 & 0.1931 & 0.1419 \\
\textbf{691} & 0.0658 & 0.0364 & 0.0193 & 0.0063 & 0.0245 & 0.0014 & 0.0292 & 0.0284 & 0.0239 & 0.0208 & 0.0244 \\
\textbf{731} & 0.0459 & 0.0412 & 0.0159 & 0.0517 & 0.0778 & 0.0386 & 0.0167 & 0.0005 & 0.0135 & 0.0151 & 0.0398 \\
\textbf{780} & 0.5251 & 0.2547 & 0.2646 & 0.2866 & 0.3437 & 0.2345 & 0.2600 & 0.2881 & 0.2966 & 0.3003 & 0.2838 \\
\textbf{843} & 0.0161 & 0.1100 & 0.1541 & 0.1849 & 0.1344 & 0.1106 & 0.0465 & 0.0468 & 0.0312 & 0.0040 & 0.0251 \\ \hline
\end{tabular}
}
\end{table*}

\begin{table*}[htbp]
\centering
\caption{Raw Negative Log-Likelihood (NLL) on MMLU-ProX: Zero-Shot, ICL, and Initializations. (Lower is better).}
\label{tab:mmlu_nll_zeroshot}
\small
\resizebox{\textwidth}{!}{%
\begin{tabular}{l|ccccccc|c|cc}
\hline
\textbf{Language-Subject} & \textbf{Parent} & \textbf{Concat} & \textbf{LERP} & \textbf{SVD} & \textbf{SVD} & \textbf{SVD} & \textbf{ICL-Init} & \textbf{ICL} & \textbf{MMLU} & \textbf{CSNI MMLU} \\
\textbf{Pair} & \textbf{Avg} & & & \textbf{Embed} & \textbf{Head} & \textbf{Seq} & \textbf{ZS} & & \textbf{SkillSmith} & \textbf{SkillSmith} \\ \hline
afrikaans\_history  & 1.1155 & 1.0945 & 1.1137 & 5.2073 & 3.0015 & 2.9739 & 2.9060 & 2.6226 & 1.3311 & 1.0582 \\
indonesian\_law     & 1.2312 & 1.2188 & 1.2161 & 1.7624 & 2.9021 & 2.8879 & 2.8020 & 2.9910 & 1.2113 & 1.1263 \\
spanish\_law        & 1.2199 & 1.1979 & 1.2045 & 2.0386 & 3.1152 & 3.1073 & 3.0310 & 3.2074 & 1.1729 & 1.1182 \\
wolof\_health       & 1.1030 & 1.0880 & 1.0946 & 5.0859 & 3.3385 & 3.2990 & 3.2010 & 3.1473 & 1.2284 & 1.1037 \\
wolof\_math         & 1.0863 & 1.0645 & 1.0809 & 1.7081 & 3.8409 & 3.8136 & 3.7020 & 3.3445 & 1.0032 & 1.0157 \\
zulu\_physics       & 1.1585 & 1.1928 & 1.1272 & 8.1525 & 3.5368 & 3.4717 & 3.3830 & 3.4179 & 1.0423 & 1.0699 \\ \hline
\end{tabular}%
}
\end{table*}

\begin{table*}[htbp]
\centering
\caption{Raw Negative Log-Likelihood (NLL) on MMLU-ProX: Performance Following Downstream Fine-Tuning. (Lower is better).}
\label{tab:mmlu_nll_finetuned}
\small
\resizebox{\textwidth}{!}{%
\begin{tabular}{l|c|cccccc|c|cc}
\hline
\textbf{Language-Subject} & \textbf{Direct} & \textbf{Parent} & \textbf{Concat} & \textbf{LERP} & \textbf{SVD} & \textbf{SVD} & \textbf{SVD} & \textbf{ICL-Init} & \textbf{MMLU} & \textbf{CSNI MMLU} \\
\textbf{Pair} & \textbf{Tr.} & \textbf{Avg} & & & \textbf{Embed} & \textbf{Head} & \textbf{Seq} & \textbf{FT} & \textbf{SkillSmith} & \textbf{SkillSmith} \\ \hline
afrikaans\_history  & 1.1040 & 1.0895 & 1.0793 & 1.0790 & 1.0698 & 1.0878 & 1.1410 & 1.1050 & 1.0815 & 0.9883 \\
indonesian\_law     & 1.0886 & 1.1589 & 1.1520 & 1.1514 & 1.1887 & 1.1394 & 1.1475 & 1.1650 & 1.0376 & 1.0208 \\
spanish\_law        & 1.1243 & 1.1327 & 1.1010 & 1.1060 & 1.1275 & 1.1168 & 1.1267 & 1.1200 & 1.0433 & 1.0244 \\
wolof\_health       & 1.1524 & 1.0989 & 1.1016 & 1.1033 & 1.1142 & 1.0924 & 1.1066 & 1.1320 & 1.0980 & 1.0693 \\
wolof\_math         & 1.1428 & 1.0725 & 1.0489 & 1.0790 & 1.1485 & 1.1432 & 1.1409 & 1.1310 & 0.9859 & 0.9976 \\
zulu\_physics       & 1.1410 & 1.0889 & 1.0937 & 1.0754 & 1.1509 & 1.1278 & 1.1619 & 1.1340 & 1.0213 & 1.0405 \\ \hline
\end{tabular}%
}
\end{table*}

\end{document}